%% file: main_arxiv.tex
\documentclass[letterpaper]{article}
\usepackage[preprint]{aaai2027} 
\usepackage[hyphens]{url} 
\usepackage{graphicx} 
\usepackage{natbib} 
\usepackage{caption} 
\usepackage{amsmath}
\usepackage{array}
\usepackage{booktabs}
\usepackage{makecell}
\usepackage{multirow}
\usepackage{pifont}
\usepackage{placeins}
\usepackage{tabularx}
\usepackage[table]{xcolor}
\usepackage{enumitem}

\newcolumntype{Y}{>{\centering\arraybackslash}X}
\definecolor{sectiongray}{gray}{0.93}
\definecolor{ourrow}{gray}{0.96}   

\definecolor{topone}{RGB}{120,180,230}
\definecolor{toptwo}{RGB}{170,210,245}
\definecolor{topthree}{RGB}{228,242,253}

\definecolor{instancecolor}{RGB}{250,235,235}
\definecolor{eventcolor}{RGB}{235,245,235}
\definecolor{scenecolor}{RGB}{250,245,230}

\newcommand{\rankone}[1]{\cellcolor{topone}#1}
\newcommand{\ranktwo}[1]{\cellcolor{toptwo}#1}
\newcommand{\rankthree}[1]{\cellcolor{topthree}#1}

\newcommand{\taubench}{TAU-Bench}
\newcommand{\airs}{A-IRS}
\newcommand{\tscore}{TraceScore}

\newcommand{\rankbox}[2]{%
\raisebox{0pt}[\height][\depth]{%
\setlength{\fboxsep}{1pt}%
\colorbox{#1}{#2}}%
}

\newcommand{\equalmark}{\textsuperscript{*}}
\newcommand{\corrmark}{\textsuperscript{\ensuremath{\dagger}}}

\newcolumntype{L}[1]{%
  >{\raggedright\arraybackslash}p{#1}%
}

\definecolor{mygreen}{RGB}{0,170,0}
\definecolor{myred}{RGB}{220,0,0}

\newcommand{\cmark}{\textcolor{mygreen}{\ding{51}}}
\newcommand{\xmark}{\textcolor{myred}{\ding{55}}}

\begin{document}
\title{TAU-Bench: From Anomaly Instance Tracking to Fine-Grained Video Anomaly Understanding}

\author{
Kepeng Yang\equalmark\textsuperscript{\rm 1},
Dongxuan Liu\equalmark\textsuperscript{\rm 1},
Rongxin Gao\equalmark\textsuperscript{\rm 1},
Zixin Su\textsuperscript{\rm 1},
Rui Wu\textsuperscript{\rm 1},
Shuzhao Xie\textsuperscript{\rm 2},
Chenxin Li\textsuperscript{\rm 3},\\
Panwang Pan\textsuperscript{\rm 1},
Yuzhi Huang\corrmark\textsuperscript{\rm 2},
Yue Huang\corrmark\textsuperscript{\rm 1},
Jingyan Jiang\textsuperscript{\rm 4}
}

\affiliations{
\textsuperscript{\rm 1}XMU,
\textsuperscript{\rm 2}THU,
\textsuperscript{\rm 3}CUHK,
\textsuperscript{\rm 4}SZTU\\[0.45em]
\equalmark Equal contribution.
\qquad
\corrmark Corresponding author.
}

\maketitle

\begin{figure*}[t]
    \centering
    \includegraphics[width=\textwidth]{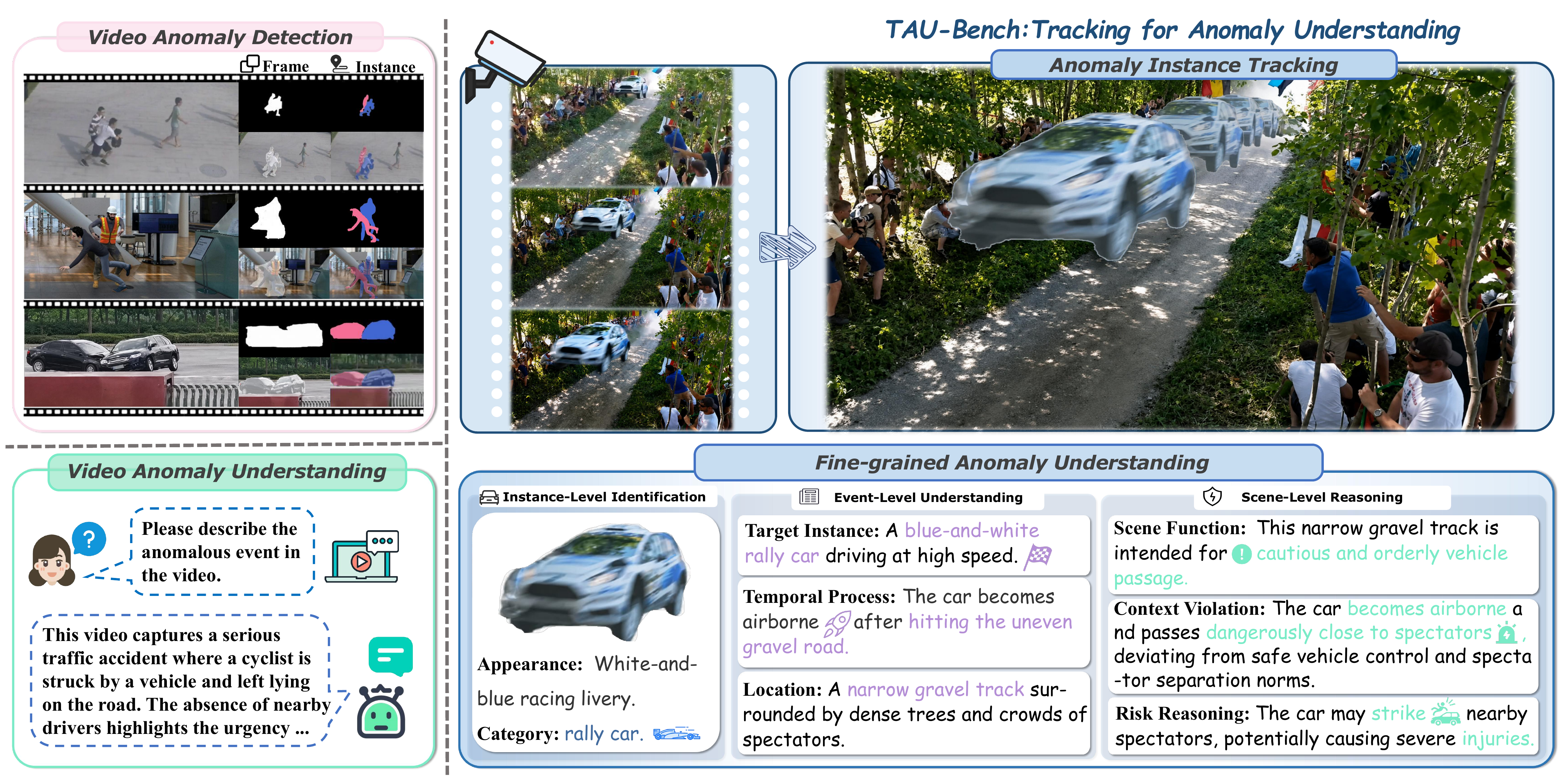}
    \caption{\textbf{Overview of TAU-Bench.}
    Most existing benchmarks provide frame-level masks or coarse video-level
    descriptions, whereas TAU-Bench associates identity-consistent
    anomaly instance tracks with fine-grained anomaly understanding.}
    \label{fig:teaser}
\end{figure*}


\begin{abstract}  Humans understand anomalous events through a coherent perceptual process in which they identify the focal instance, follow its behavior as the event unfolds, and interpret why it violates the expectations of the surrounding scene. Video anomaly understanding (VAU) seeks to endow models with a similar capability, moving beyond deciding whether a video is anomalous toward explaining how the event develops and why it matters. Although recent vision--language models (VLMs) can generate detailed and plausible anomaly descriptions, their semantic fluency does not ensure that these interpretations remain grounded in the correct anomaly instance over time. Existing benchmarks typically evaluate tracking and semantic understanding through separate protocols, leaving such instance--semantic inconsistency largely unmeasured. We therefore introduce TAU-Bench, a track-centric benchmark for jointly evaluating anomaly instance tracking and fine-grained anomaly understanding. TAU-Bench contains 1,118 videos, 1,454 tracks, and 202,438 pixel-level masks spanning 49 event and 45 scene categories, together with track-centric annotations that connect instance-level identification, event-level understanding, and scene-level reasoning. To build \textbf{TAU-Bench} at scale, we developed an automated data engine integrating anomaly suitability filtering, anomaly instance track construction, hierarchical caption annotation, and human quality control. Evaluations across representative VLM families show that models producing plausible anomaly interpretations may still fail to localize and track the correct instance reliably, revealing a persistent gap between semantic reasoning and visual grounding. These findings therefore highlight instance-grounded evaluation as an important step toward more faithful and reliable VAU systems.
\end{abstract}


\section{Introduction}
\label{sec:introduction}
Video anomaly understanding (VAU) requires perceiving how a focal anomaly instance, its behavior, and its relation to the surrounding scene evolve over time\cite{sanders2024survey,qian2024rethinking}. This dynamic perception, which centers on a persistent anomaly instance, allows humans to identify what makes an event abnormal, track how it unfolds, and interpret why it matters within a specific context\cite{du2024uncovering,tang2024hawk}. Although recent vision--language models (VLMs) can generate detailed anomaly descriptions and infer possible consequences, whether they can maintain such instance grounding as an event evolves remains underexplored. This motivates us to ask: \textit{Can current VLMs keep their interpretations anchored to the correct anomaly instance over time?}

Existing VAU methods, which have progressively moved beyond binary anomaly detection, now support temporal localization, open-ended event description, and context-dependent reasoning~\cite{zanella2024harnessing,ye2025vera,zhang2025holmes,gao2025vagu,tang2024hawk,huang2025exvad,du2024uncovering,yang2024follow,zhu2025vaur1,yu2025cuebench}, while parallel efforts in pixel-level localization, anomalous-object tracking, and target-centric understanding have made anomaly analysis increasingly instance-aware~\cite{huang2025tao,gu2026tvau,zhou2026targetvau}. Despite these complementary advances, the corresponding capabilities are generally developed as separate tasks, which leaves it unclear whether a model's localization and semantic interpretation remain anchored to the same anomaly instance throughout an event.

Current VAU benchmarks have expanded their annotations from coarse anomaly labels to dense event descriptions, temporal grounding, open-ended question answering, and multi-level contextual reasoning~\cite{yuan2024surveillance,tang2024hawk,du2024uncovering,zhou2024human,zhang2025holmes,gao2025vagu,liu2025surveillancevqa,pereira2026finevau,yu2025cuebench}. Nevertheless, they still reflect the same separation at the method level because tracking and semantic understanding are evaluated through distinct task-specific protocols. As summarized in Tab.~\ref{tab:benchmark_comparison}, traditional VAD datasets primarily support anomaly detection, whereas recent VAU benchmarks add temporal or semantic supervision but do not jointly provide identity-consistent tracks, pixel-level masks, and instance, event and scene-level understanding. Consequently, current protocols cannot determine whether a model remains grounded in the instance it describes.

To bridge this methodological and evaluative gap, we introduce \textbf{TAU-Bench} (\textbf{T}racking for \textbf{A}nomaly \textbf{U}nderstanding), a track-centric benchmark that unifies anomaly instance tracking with fine-grained anomaly understanding. As illustrated in Fig.~\ref{fig:teaser}, TAU-Bench assigns each anomaly instance an identity-consistent pixel-level track spanning the anomalous interval, which serves as a shared visual anchor that aligns spatial, temporal, and semantic supervision. Building on this representation, the benchmark evaluates two complementary capability families, with \emph{Anomaly Instance Tracking} covering video-level anomaly detection, temporal anomaly localization, and anomaly instance reasoning segmentation (A-IRS), while \emph{Fine-grained Anomaly Understanding} assesses instance-level identification, event-level understanding, and scene-level reasoning. The resulting benchmark comprises 1,118 videos, 1,454 anomaly instance tracks, and 202,438 pixel-level masks across 49 event categories and 45 scene categories, with its supervision constructed using a scalable data engine that filters annotation-suitable videos, generates identity-consistent track proposals, and produces hierarchical captions, after which human verification ensures cross-modal consistency. Evaluations across representative VLM families reveal a clear gap between plausible anomaly interpretation and reliable instance grounding, underscoring the need to evaluate tracking and understanding jointly.

Our main contributions are summarized as follows:
\begin{itemize}
\item We formulate fine-grained VAU as a coupled tracking and understanding problem, establishing a track-centric evaluation framework that measures whether spatial, temporal, and semantic predictions remain grounded in the same anomaly instance to assess cross-task factual consistency.
\item We construct \textbf{TAU-Bench}, a large-scale benchmark comprising 1,118 videos, 1,454 identity-consistent tracks, and 202,438 pixel-level masks across 49 event and 45 scene categories. Notably, this unified  benchmark supports complementary tasks spanning video-level anomaly detection, temporal localization, pixel-level instance tracking, and hierarchical understanding of anomalies.
\item We develop a scalable data construction engine that integrates filtering for anomaly videos, the construction and verification of identity-consistent tracks, and hierarchical captioning, thereby enabling the large-scale generation of multimodal data for downstream model training.

\end{itemize}


\input{Figures/bench_comparison}

\begin{figure*}[!t]
    \centering
    \includegraphics[width=\textwidth]{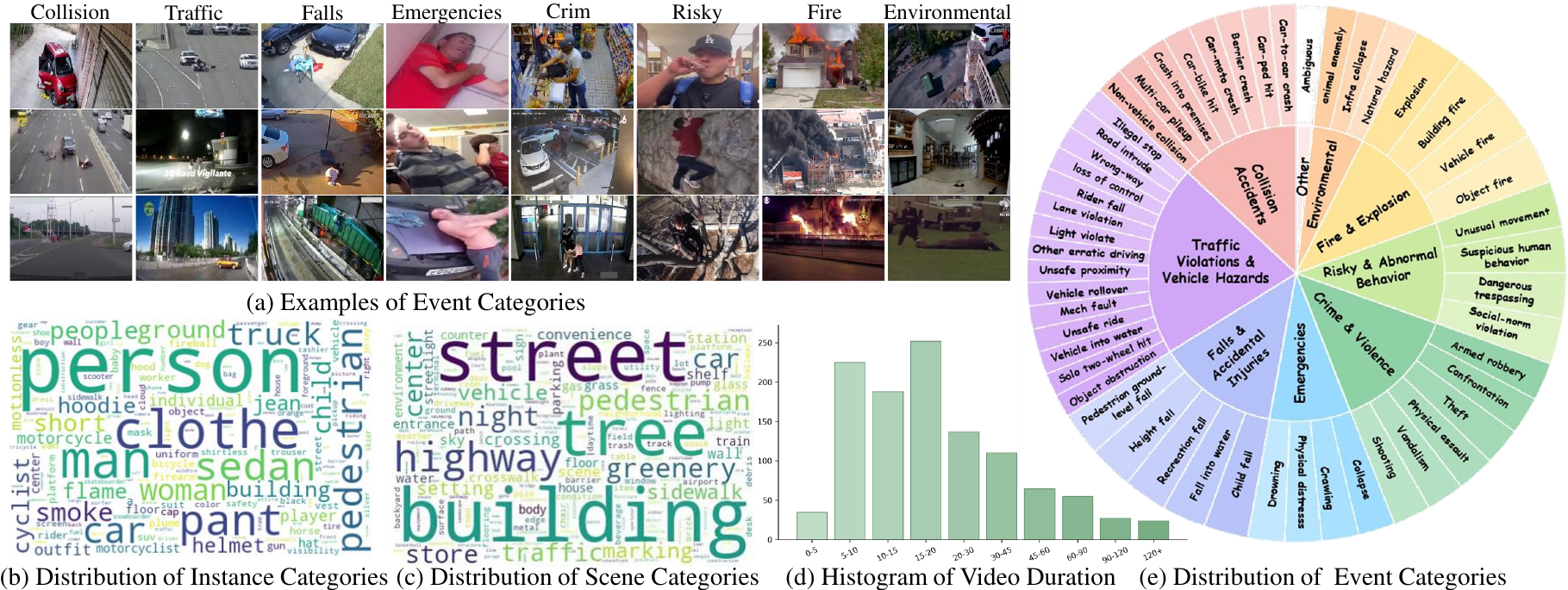}
    \caption{Data statistics of TAU-Bench, including anomaly event examples and distributions over instance, event, scene categories. 
    }
    \label{fig:data_statistics}
\end{figure*}


\begin{figure*}[t]
    \centering
    \includegraphics[width=0.98\textwidth]{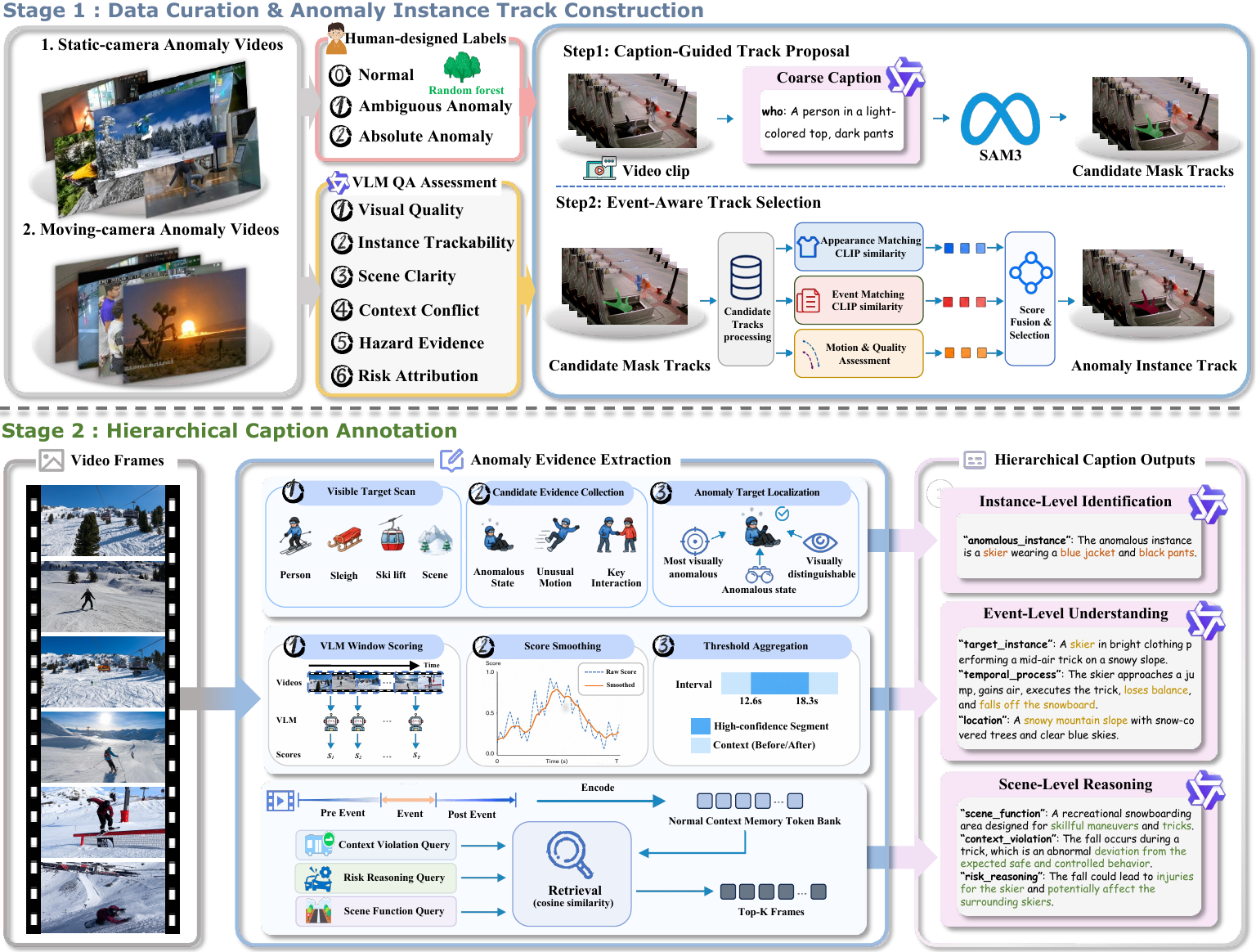}
    \caption{Data construction engine of TAU-Bench. Stage 1 curates static- and dynamic-camera anomaly videos and constructs identity-consistent anomaly instance tracks. Stage 2 performs Anomaly Evidence Extraction (AEE) to generate progressive instance-level, event-level, and scene-level captions }
    \label{fig:pipeline}
\end{figure*}



\section{Related Work}
\subsection{VAU Methods}

Recent VAU methods extend anomaly analysis beyond binary detection to temporal localization, open-ended description, and contextual reasoning. HAWK strengthens motion aware anomaly interpretation, Holmes-VAU extracts anomaly relevant moments from long videos, while CUVA, VAU-R1, and Cue-R1 explore causal or context dependent reasoning~\cite{tang2024hawk,zhang2025holmes,du2024uncovering,zhu2025vaur1,yu2025cuebench}. VAGU couples coarse anomaly localization with semantic interpretation and temporal-boundary refinement~\cite{gao2025vagu}. Concurrently, TAO, T-VAU, and TargetVAU advance pixel-level tracking, localized explanation, and target-centric understanding~\cite{huang2025tao,gu2026tvau,zhou2026targetvau}. However, these semantic and spatial capabilities are developed under separate objectives. Consequently, models may generate plausible descriptions without anchoring their reasoning to the same anomaly instance throughout the event.

\subsection{VAU Benchmarks}

Traditional VAD datasets provide video or frame-level labels for anomaly detection and localization. Recent VAU benchmarks enrich this with event descriptions, temporal boundaries, open-ended question answering, and structured reasoning. UCA and HAWK support anomaly captioning and interaction; CUVA annotates occurrences, causes, and consequences; Holmes-VAU provides multi-scale instructions; VAGU evaluates temporal grounding and semantic understanding; while SurveillanceVQA, FineVAU, and CueBench examine comprehensive, fine-grained, or context dependent reasoning~\cite{yuan2024surveillance,tang2024hawk,du2024uncovering,zhang2025holmes,gao2025vagu,liu2025surveillancevqa,pereira2026finevau,yu2025cuebench}. However, these annotations target videos, intervals, or text, while some benchmarks provide masks or tracks lacking multi-level semantics. Thus, existing protocols cannot verify identity-consistent localization alongside instance, event and scene-level understanding.

\section{TAU-Bench}
\label{sec:tau_bench}

\subsection{Overview}
We present \textbf{TAU-Bench}, a unified benchmark for evaluating VLMs in anomaly instance tracking and fine-grained video anomaly understanding. Compared with existing datasets and benchmarks in Tab.~\ref{tab:benchmark_comparison}, TAU-Bench contains 1,118 videos, 1,454 identity-consistent anomaly tracks, and 202,438 pixel-level masks across 49 event and 45 scene categories. It defines two task families: \emph{Anomaly Instance Tracking}, covering anomaly detection, temporal localization, and anomaly instance reasoning segmentation, and \emph{Fine-Grained Anomaly Understanding}, covering instance-level identification, event-level understanding, and scene-level reasoning. Fig.~\ref{fig:data_statistics} summarizes the data statistics. Inspired by recent efficient modular pipelines that compose pretrained foundation models without training~\cite{wen2025dynamicverse, huang2026thinkingdynamicsmultimodallarge, huang2026robostreamweavingspatiotemporalreasoning, gao2026dyntrace}, our data engine separately constructs pixel-level instance tracks and hierarchical captions, then aligns them to the same focal anomaly identity through human verification.

\subsection{Data Construction Engine}

\paragraph{Data Curation.}We collect two types of candidate data, including static-camera anomaly videos: ShanghaiTech~\cite{luo2017revisit},
UBnormal~\cite{acsintoae2022ubnormal}, and moving-camera anomaly videos:
CueBench~\cite{yu2025cuebench}, MSAD~\cite{zhu2024advancing}, NWPU
Campus~\cite{cao2023comprehensive},  and TAD~\cite{xu2025tad}. Together,
these sources cover campus environments, surveillance footage, driving
scenes, and open-world Internet videos. We perform \emph{VLM QA Assessment} along six dimensions: visual quality, instance trackability, scene clarity, context conflict, hazard evidence, and risk attribution. A subset is further assigned \emph{Human-designed Labels}: 0 for \emph{Normal}, 1 for \emph{Ambiguous Anomaly}, and 2 for \emph{Absolute Anomaly}. These labels supervise a Random Forest filter. The pipeline reduces 4,448 candidates to 2,792 automatically selected samples, from which 1,118 videos are retained after human verification. Further details are provided in the supplementary material.

\input{Figures/Experiments/main_caption}

\vspace{-3pt}
\paragraph{Anomaly Instance Track Construction.}
Track construction comprises \emph{Caption-Guided Track Proposal} and \emph{Event-Aware Track Selection}. For each retained video, a pretrained VLM generates a coarse \textit{who}--\textit{what} caption: \textit{who} identifies the visible anomaly instance and \textit{what} summarises its anomalous action. The \textit{who} description prompts SAM3~\cite{carion2025sam3segmentconcepts} to generate temporally linked candidate mask tracks. For each candidate track $\tau$, the module samples masked crops that isolate the target instance and context crops that preserve surrounding event information. \emph{Appearance Matching} and \emph{Event Matching} compute $S_{\mathrm{app}}$ and $S_{\mathrm{evt}}$, respectively, using CLIP~\cite{clip} similarity between the sampled crops and the corresponding \textit{who} and \textit{who}--\textit{what} descriptions. From each track's masks, we derive per-frame bounding boxes, mask areas, and frame indices to compute two additional scores: $S_{\mathrm{mot}}$ reflects motion plausibility based on box displacement, trajectory length, and shape changes, while $S_{\mathrm{qual}}$ reflects whether the masks are valid, temporally continuous, and reasonable in scale. These scores are fused as $S(\tau)=\sum_{k\in\mathcal{K}}\lambda_k S_k(\tau)$, where $\mathcal{K}={\mathrm{app},\mathrm{evt},\mathrm{mot},\mathrm{qual}}$, and the highest-scoring candidate is selected as the anomaly instance track.

\vspace{-5pt}
\paragraph{Hierarchical Caption Annotation.}

Motivated by progressive human anomaly perception, we formulate hierarchical caption annotation as \emph{Anomaly Evidence Extraction} (AEE), which constructs three independently supervised yet semantically progressive levels linked by the same anomaly instance: instance-level identification, event-level understanding, and scene-level reasoning. The resulting annotation is defined as $\mathcal{H}(V)=\{c_{\mathrm{ins}},c_{\mathrm{evt}},c_{\mathrm{scn}}\}$.
For instance-level identification, the complete video is provided to the VLM, which scans visible targets, collects anomaly evidence of candidates, and localizes the anomaly instance. The VLM generates $c_{\mathrm{ins}}$, a concise description of the visible category and appearance of the target. For event-level understanding, the video is divided into overlapping windows, and the VLM assigns an anomaly score to each window. Score smoothing and threshold aggregation are then applied to localize the event interval. The localized segment, together with a short temporal context, is provided to the VLM to generate $c_{\mathrm{evt}}$, which comprises the \textit{target instance}, \textit{temporal process}, and \textit{location}. For scene-level reasoning, frames outside the event interval are encoded into a visual memory bank. The three fields of $c_{\mathrm{evt}}$ are used to formulate queries concerning scene function, context violation and risk reasoning. The query embeddings are aligned with the visual embeddings to retrieve keyframes. Together with the event segment, these keyframes guide the VLM to generate $c_{\mathrm{scn}}$, which includes \textit{scene function}, \textit{context violation}, and \textit{risk reasoning}.

\vspace{-5pt}
\paragraph{Human Quality Control.} Automatically generated videos, tracks, intervals, and captions are treated as annotation proposals. Annotators verify anomaly visibility, instance uniqueness, event completeness, and contextual interpretability; inspect each track frame by frame to correct identity switches, fragmentation, missing masks, false observations, and inaccurate boundaries; and evaluate the instance-, event-, and scene-level descriptions against the raw video, annotated interval, and corrected track. Samples with unresolved ambiguity, irreparable tracks, incomplete events, or unsupported causal or risk statements are discarded. The supplementary material provides details on secondary review, disagreement resolution, correction rates, and the complete annotation guidelines.

\subsection{Evaluation Framework}
\label{sec:evaluation}

\paragraph{Task Definition}
\input{Figures/Experiments/recall_tiou_table}
\vspace{-3pt}

\input{Figures/Experiments/reasoning_segmentation}

\input{Figures/Experiments/ablation_table}

Let $V=\{I_t\}_{t=1}^{T}$ denote input video, and let $q$ denote task-specific prompts. TAU-Bench evaluates two complementary capabilities: \textit{Anomaly Instance Tracking} and \textit{Fine-Grained Anomaly Understanding}.

\textit{Anomaly Instance Tracking.}
This capability comprises three tasks with progressively finer outputs. \emph{(i) Video-level Anomaly Detection} takes $V$ to predict whether the video contains an anomaly. \emph{(ii) Temporal Anomaly localization} takes $V$ to predict the temporal interval covering the complete anomalous process. 
\emph{(iii) Anomaly Instance Reasoning Segmentation (A-IRS)} takes $V$ and optionally an anomaly description $q_{\mathrm{ano}}$ as input, outputting an identity-consistent mask track of the target anomaly instance throughout the anomalous interval.
A-IRS is evaluated under three settings: \emph{Generated Caption}, which uses the model's own anomaly description to evaluate semantic to spatial transfer; \emph{GT Caption}, which isolates visual grounding under correct semantic guidance; and \emph{Direct Prompt}, which evaluates end-to-end reasoning segmentation without an explicit semantic intermediate. Their comparison distinguishes errors arising from semantic guidance from those arising from visual grounding.

\textit{Fine-Grained Anomaly Understanding.}
This ability is evaluated via structured caption generation from videos. Given $V$ and a task-specific prompt, the model generates seven anomaly-grounded semantic fields: instance description at the instance level; target instance, temporal process, and location at the event level; and scene function, context violation, and risk reasoning at the scene level.

\paragraph{Metric Design}
TAU-Bench uses task-aligned metrics per output type. All models are evaluated zero-shot via task-specific prompts and deterministic parsing. Supplementary materials detail prompt templates and implementation.

\textit{Tracking Metrics.}
We report anomaly-class Recall for video-level anomaly detection, and temporal Intersection-over-Union (tIoU) between predicted and reference intervals for temporal localization. For A-IRS, we evaluate predicted mask tracks within annotated anomaly intervals using region similarity $\mathcal{J}$, boundary accuracy $\mathcal{F}$, and their mean $\mathcal{J}\&\mathcal{F}$.

\textit{Understanding Metrics.}
Conventional lexical metrics capture surface-word overlap, but they are insufficient for evaluating the semantic fidelity of free-form anomaly descriptions. We therefore introduce TraceScore, a VLM-judge-based metric with lexical support and contradiction control. 
For each sample $i$ and semantic field $d$, given the reference caption $c_{i,d}$ and prediction $\hat{c}_{i,d}$, a fixed VLM judge assigns a normalized semantic similarity score $s^{\mathrm{sem}}_{i,d}$, while METEOR~\cite{banerjee-lavie-2005-meteor} provides a lexical support score $s^{\mathrm{lex}}_{i,d}$.
We define $p_{i,d}=\max(0, s^{\mathrm{sem}}_{i,d}-s^{\mathrm{lex}}_{i,d})$ to measure judge-assigned semantic credit that is weakly supported by lexical evidence, and use a contradiction gate $C_{i,d}\in\{0,1\}$ to suppress predictions incompatible with the reference instance, event, or scene interpretation. The final score is computed as $\mathrm{TS}_{i,d}=C_{i,d}s^{\mathrm{sem}}_{i,d}(1-\lambda p_{i,d})$, where $\lambda$ controls the lexical-support penalty. TraceScore thus rewards semantically faithful descriptions while discounting contradictory or weakly grounded matches.

\input{Figures/Experiments/case_study}

\section{Evaluation on TAU-Bench}

We evaluate TAU-Bench around a human-like progressive view of anomaly understanding: identifying the anomaly instance, following the event, and reasoning about scene-level violation and risk. We first assess fine-grained understanding across the instance, event, and scene hierarchy, then evaluate anomaly instance tracking through video-level detection, temporal localization, and A-IRS. The A-IRS settings further diagnose how semantic descriptions interact with visual grounding under different guidance conditions. Ablation studies and case study validate the annotation design and further illustrate the observed cross-task behavior. All models are evaluated zero-shot with default instruction templates.

\subsection{Fine-Grained Anomaly Understanding}

Tab.~\ref{tab:caption_results} evaluates whether VLMs can follow the fine-grained semantic hierarchy of TAU-Bench. This hierarchy reflects the progressive way humans understand anomalies. The results show fragmented strengths across levels, with no model consistently dominating all semantic fields.
Qwen3-VL-8B-SAMTok performs best on instance-level identification, suggesting the benefit of segmentation-oriented grounding. VideoChat-R1.5 and InternVL3.5 are stronger in event-level understanding, while InternVL3.5 performs well on scene-level reasoning. These patterns indicate that current VLMs may recognize the anomalous object, describe the event, or infer contextual risk, but they do not reliably maintain a coherent instance, event and scene interpretation. TAU-Bench therefore exposes fine-grained semantic failures hidden by coarse anomaly labels or single free-form captions.

\vspace{-2mm}
\subsection{Anomaly Instance Tracking}
Tab.~\ref{tab:recall_tiou_table} reports the performance on video-level anomaly detection and temporal anomaly localization. All models achieve high anomaly recall, ranging from 89.28 to 98.29, whereas temporal localization remains unsaturated, with the best tIoU reaching 79.20. Moreover, the models achieving the best recall and tIoU differ, indicating that detecting anomaly presence and recovering its complete temporal extent require distinct capabilities.
Tab.~\ref{tab:JF_table} evaluates A-IRS under Generated Caption, GT Caption, and Direct Prompt. GT Caption improves $\mathcal{J}\&\mathcal{F}$ for all six models, with an average gain of 11.82 points over Generated Caption, showing that imperfect self-generated semantics are a major source of downstream tracking errors. Generated Caption outperforms Direct Prompt for only half of the models, indicating that an explicit semantic intermediate facilitates grounding only when the generated description provides reliable instance guidance. Meanwhile, the remaining gap under GT Caption shows that correct semantics alone does not ensure accurate visual grounding.

\subsection{Ablation Studies}

We conduct ablation studies on the proposed AEE strategies for hierarchical anomaly understanding. Tab.~\ref{tab:ablation} demonstrates the effectiveness of AEE across the three levels. At instance-level, AEE improves AI from 41.32 to 42.56, indicating better focal anomaly instance extraction. At event-level, it improves tIoU by 9.83 points and benefits TI, TP, and Loc, showing that localized event evidence yields cleaner and more accurate structured event captions. At scene-level, the gains on SF, CV, and RR confirm that event-conditioned scene evidence supports scene function, context violation, and risk reasoning more effectively. Overall, the ablation supports our design of independent yet progressive captions grounded from instance to event and scene, validating the effectiveness of the proposed hierarchical annotation strategy.

\subsection{Case Study}
Fig.~\ref{fig:case_study} illustrates how TAU-Bench connects fine-grained anomaly understanding with anomalous instance tracking. In this gas-station accident, L1 requires the model to identify the anomalous instance, i.e., the black car, without relying on full event semantics. L2 then binds this instance to the event by specifying the target instance, its collision with the fuel pump, and the gas station location. Based on these fields, L3 further infers the violated scene function and the potential fuel leak/fire risks. The tracking panel shows that this semantic chain must correspond to the anomaly instance track rather than nearby distractors, highlighting TAU-Bench's focus on identity-consistent anomaly understanding across tasks.
\FloatBarrier
\section{Conclusion}

We introduced \textbf{TAU-Bench}, a track-centric benchmark for jointly evaluating fine-grained anomaly understanding and anomaly instance tracking. TAU-Bench provides 1,118 videos, 1,454 identity-consistent anomaly instance tracks, and 202,438 pixel-level masks, together with hierarchical instance-level, event-level, and scene-level captions. To build TAU-Bench at scale, we developed an automatic data construction engine integrating anomaly suitability filtering, anomaly instance track construction, hierarchical caption annotation, and human quality control. Experiments across representative VLMs suggest that current models show fragmented strengths across semantic fields and benefit from reliable semantic guidance in A-IRS, but still leave room for more coherent and comprehensive instance-grounded anomaly understanding. TAU-Bench thus provides a unified testbed for systematically studying how anomaly semantics and visual grounding interact over time.
\bibliography{aaai2027}

\input{supplementary_arxiv_content}

\end{document}

%% file: Figures/bench_comparison.tex
\newcolumntype{C}[1]{>{\centering\arraybackslash}p{#1}}
\begin{table*}[t]
\centering
\begingroup

\small
\setlength{\tabcolsep}{1.0pt}       
\renewcommand{\arraystretch}{0.96}  

\small
\setlength{\tabcolsep}{1.35pt}
\renewcommand{\arraystretch}{0.96}

\begin{tabularx}{\textwidth}{
@{}
>{\raggedright\arraybackslash}X|
*{3}{c}|
*{5}{c}|
*{3}{c}|
*{3}{C{0.038\textwidth}}
@{}
}

\toprule

\multirow{2}{*}{\textbf{Benchmark}}
& \multicolumn{3}{c|}{\textsc{Data Scale}}
& \multicolumn{5}{c|}{\textsc{Anomaly Supervision}}
& \multicolumn{3}{c|}{\textsc{Tracking}}
& \multicolumn{3}{c}{\textsc{Understanding}}
\\

\cmidrule(lr){2-4}
\cmidrule(lr){5-9}
\cmidrule(lr){10-12}
\cmidrule(lr){13-15}

& \textbf{Domain}
& \textbf{Length}
& \textbf{Videos}
& \textbf{Inst}
& \textbf{Evt}
& \textbf{Scn}
& \textbf{Track}
& \textbf{Pixel}
& \textbf{VAD}
& \textbf{T-Loc}
& \textbf{A-IRS}
& \textbf{I}
& \textbf{E}
& \textbf{S}
\\

\midrule

\rowcolor{sectiongray}
\multicolumn{15}{c}{%
    \textit{\textbf{Traditional Video Anomaly Detection Datasets}}
}
\\
\midrule

Subway Entrance \cite{adam2008robust}
& Ped         & 1.5h & 1
& -- & 5 & 1 & -- & --
& \cmark & \cmark & \xmark
& \xmark & \xmark & \xmark
\\

UCSD Ped1 \cite{wang2010anomaly}
& Ped & 0.1h & 36
& 61 & 5 & 1 & -- & 2,389
& \cmark & \cmark & \xmark
& \xmark & \xmark & \xmark
\\

UCSD Ped2 \cite{wang2010anomaly}
& Ped & 0.05h & 12
& 21 & 5 & 1 & -- & 842
& \cmark & \cmark & \xmark
& \xmark & \xmark & \xmark
\\

ShanghaiTech \cite{luo2017revisit}
& Ped & 1.9h & 107
& 158 & 13 & 1 & -- & 8,907
& \cmark & \cmark & \xmark
& \xmark & \xmark & \xmark
\\

\mbox{UCF-Crime\cite{sultani2018realworld}}
& Crime & 64h & 950
& -- & 13 & 13 & -- & --
& \cmark & \cmark & \xmark
& \xmark & \xmark & \xmark
\\

XD-Violence \cite{wu2020notonly}
& Crime & 110h & 2,405
& -- & 6 & -- & -- & --
& \cmark & \cmark & \xmark
& \xmark & \xmark & \xmark
\\

UBnormal \cite{acsintoae2022ubnormal}
& Ped & 0.85h & 211
& 660 & 22 & 29 & -- & 58,983
& \cmark & \cmark & \xmark
& \xmark & \xmark & \xmark
\\

NWPU Campus \cite{cao2023comprehensive}
& Ped & 0.73h & 124
& -- & 28 & 43 & -- & --
& \cmark & \cmark & \xmark
& \xmark & \xmark & \xmark
\\

TAD \cite{TAD}
& Traffic & 12.5h & 250
& -- & 7 & -- & -- & --
& \cmark & \cmark & \xmark
& \xmark & \xmark & \xmark
\\

MSAD \cite{zhu2024advancing}
& Multi & 1.38h & 240
& -- & 55 & 14 & -- & --
& \cmark & \cmark & \xmark
& \xmark & \xmark & \xmark
\\

\midrule

\rowcolor{sectiongray}
\multicolumn{15}{c}{%
    \textit{\textbf{Video Anomaly Understanding Benchmarks}}
}
\\
\midrule

CUVA \cite{du2024uncovering}
& Multi & 32.46h & 1,000
& -- & 42 & -- & -- & --
& \cmark & \cmark & \xmark
& \xmark & \cmark & \cmark
\\

HAWK \cite{tang2024hawk}
& Multi & 142.5h & 7,898
& -- & -- & -- & -- & --
& \cmark & \cmark & \xmark
& \xmark & \cmark & \xmark
\\

HIVAU-70K \cite{zhang2025holmes}
& Multi & 176.2h & 2,781
& -- & 13 & -- & -- & --
& \cmark & \cmark & \xmark
& \xmark & \cmark & \xmark
\\

CueBench \cite{yu2025cuebench}
& Multi & 26.0h & 2,950
& 1,249 & 32 & 174 & -- & --
& \cmark & \cmark & \xmark
& \cmark & \cmark & \xmark
\\

\rowcolor{blue!5}
\textbf{TAU-Bench (Ours)}
& Multi
& 8.25h
& 1,118
& 1,454
& 49
& 45
& 1,454
& 202,438
& \textbf{\cmark}
& \textbf{\cmark}
& \textbf{\cmark}
& \textbf{\cmark}
& \textbf{\cmark}
& \textbf{\cmark}
\\

\bottomrule
\end{tabularx}

\endgroup
\caption{
Comparison with video anomaly detection and understanding benchmarks. 
\textit{Inst}/\textit{Evt}/\textit{Scn} denote annotated instances, anomaly event categories, and scene categories; 
\textit{Track}/\textit{Pixel} denote instance tracks and pixel-level masks. 
\textit{VAD}/\textit{T-Loc}/\textit{A-IRS} denote video anomaly detection, temporal localization, and anomaly instance reasoning segmentation; 
\textit{I}/\textit{E}/\textit{S} denote instance-, event-, and scene-level anomaly captioning.
}
\label{tab:benchmark_comparison}
\end{table*}

%% file: Figures/Experiments/main_caption.tex

\begin{table*}[!t]
\centering
{\small
\setlength{\tabcolsep}{1mm}
\renewcommand{\arraystretch}{0.9}

\begin{tabularx}{\textwidth}{@{}l c *{6}{Y}@{}}
\toprule
\textbf{Methods}
& \multicolumn{1}{c}{
    \shortstack{
        \textbf{Instance-Level}\\
        \textbf{Identification}
    }
}
& \multicolumn{3}{c}{
    \shortstack{
        \textbf{Event-Level}\\
        \textbf{Understanding}
    }
}
& \multicolumn{3}{c}{
    \shortstack{
        \textbf{Scene-Level}\\
        \textbf{Reasoning}
    }
} \\

\cmidrule(lr){2-2}
\cmidrule(lr){3-5}
\cmidrule(lr){6-8}

& \textbf{AI}
& \textbf{TI}
& \textbf{TP}
& \textbf{Loc}
& \textbf{SF}
& \textbf{CV}
& \textbf{RR} \\

\midrule
\rowcolor{sectiongray}
\multicolumn{8}{c}{\textit{General-Purpose VLMs}} \\
\midrule


Qwen3.5-9B~\cite{qwen3.5}
& 42.56
& 38.49
& 37.45
& \rankthree{63.72}
& 60.65
& 41.92
& 48.01 \\

InternVL3-9B~\cite{internvl3}
& 36.90
& 33.29
& \rankthree{38.18}
& 41.39
& 59.98
& \rankthree{64.67}
& 59.18 \\

InternVL3-14B~\cite{internvl3}
& 39.31
& 34.68
& \ranktwo{38.61}
& 48.25
& \rankthree{61.23}
& \ranktwo{65.08}
& \ranktwo{61.57} \\

InternVL3.5-8B~\cite{internvl3_5}
& 42.62
& 33.09
& 30.76
& 52.98
& 52.76
& 36.92
& 48.34 \\

InternVL3.5-14B~\cite{internvl3_5}
& 42.84
& 39.09
& \rankone{38.72}
& 60.13
& \ranktwo{61.75}
& \rankone{66.11}
& \rankone{63.72} \\

\midrule
\rowcolor{sectiongray}
\multicolumn{8}{c}{\textit{Video-Reasoning VLMs}} \\
\midrule

Video-R1-7B~\cite{Video-R1}
& 29.98
& 20.06
& 23.46
& 40.07
& 55.90
& 59.19
& 58.54 \\

VideoChat-R1-7B~\cite{VideoChat-R1}
& 44.46
& \rankthree{39.96}
& 36.62
& 63.38
& 58.92
& 44.32
& 51.71 \\

VideoChat-R1.5-7B~\cite{VideoChat-R1.5}
& \ranktwo{44.92}
& \rankone{41.99}
& 36.89
& \ranktwo{64.03}
& 59.76
& 46.84
& 53.00 \\

VideoLLaMA3~\cite{VideoLLaMA3}
& 37.91
& 28.14
& 28.17
& 42.69
& 50.20
& 61.70
& \rankthree{61.38} \\

ReWatch-R1~\cite{ReWatch-R1}
& 31.02
& 18.00
& 12.12
& 27.17
& 45.61
& 38.00
& 47.99 \\

\midrule
\rowcolor{sectiongray}
\multicolumn{8}{c}{\textit{Reasoning-Segmentation VLMs}} \\
\midrule

Sa2VA-InternVL3-8B~\cite{sa2va}
& 31.11
& 23.88
& 14.92
& 24.17
& 42.91
& 49.83
& 55.11 \\

Sa2VA-InternVL3-14B~\cite{sa2va}
& 31.57
& 28.03
& 18.39
& 44.33
& 55.61
& 61.41
& 59.31 \\

UniPixel-3B~\cite{UniPixel}
& 16.67
& 21.83
& 15.22
& 40.79
& 40.68
& 38.20
& 47.67 \\

UniPixel-7B~\cite{UniPixel}
& 25.38
& 24.72
& 21.09
& 44.48
& 49.88
& 51.21
& 54.69 \\

VideoGLaMM~\cite{VideoGLaMM}
& 23.53
& 12.04
& 10.86
& 23.32
& 32.05
& 40.42
& 37.92 \\

Veason-R1~\cite{veason}
& 38.87
& 29.95
& 28.94
& 43.78
& 55.06
& 50.31
& 56.53 \\

Qwen3-VL-8B-SAMTok~\cite{samtok}
& \rankone{49.33}
& 37.60
& 33.27
& 44.10
& 54.23
& 54.57
& 50.68 \\

\midrule
\rowcolor{sectiongray}
\multicolumn{8}{c}{\textit{VAU-Specific VLMs}} \\
\midrule

\shortstack[l]{HolmesVAU-2B}~\cite{zhang2025holmes}
& 39.88
& 22.88
& 20.09
& 27.57
& 22.98
& 15.83
& 22.51 \\

\shortstack[l]{Vad-R1}~\cite{Vad-R1}
& \rankthree{44.62}
& \ranktwo{41.00}
& 30.27
& \rankone{72.37}
& \rankone{66.65}
& 44.33
& 51.88 \\

\shortstack[l]{Cue-R1}~\cite{yu2025cuebench}
& 41.08
& 30.39
& 25.91
& 53.74
& 51.81
& 35.55
& 46.73 \\

\bottomrule

\end{tabularx}
}

\caption{
Performance comparison on fine-grained anomaly understanding. AI denotes anomalous instance identification; TI, TP, and Loc denote target instance, temporal process, and location; SF, CV, and RR denote scene function, context violation, and risk reasoning, respectively. 
Top three performers in each category are highlighted from 
\rankbox{topone}{Dark} (highest) to 
\rankbox{topthree}{Light} (third highest).
}
\label{tab:caption_results}
\end{table*}

%% file: Figures/Experiments/recall_tiou_table.tex
\begin{table}[t]
\centering
\small
\setlength{\tabcolsep}{1.5mm}
\renewcommand{\arraystretch}{0.9    }


\begin{tabular*}{\columnwidth}{@{\extracolsep{\fill}}lcc@{}}
\toprule

\textbf{Method}
& \textbf{Recall}
& \textbf{tIoU} \\

\midrule

Qwen3.5-VL-9B~\cite{qwen3.5}
& 89.28
& 72.28 \\

InternVL3.5-14B~\cite{internvl3_5}
& \underline{97.41}
& \textbf{79.20} \\

VideoChat-R1.5-7B~\cite{VideoChat-R1.5}
& 93.82
& 69.34 \\

Qwen3-VL-8B-SAMTok~\cite{samtok}
& 93.10
& \underline{78.54} \\

Vad-R1~\cite{Vad-R1}
& 96.31
& 69.65 \\

Cue-R1~\cite{yu2025cuebench}
& \textbf{98.29}
& 76.40 \\

\bottomrule
\end{tabular*}

\caption{
Performance comparison on anomaly instance tracking. Recall evaluates video-level anomaly identification, while tIoU measures temporal localization accuracy.
}
\label{tab:recall_tiou_table}
\vspace{-5mm}
\end{table}

%% file: Figures/Experiments/reasoning_segmentation.tex
\begin{table*}[!t]
\centering
\small
\setlength{\tabcolsep}{3.2pt}
\renewcommand{\arraystretch}{0.9}

\begin{tabularx}{\textwidth}{@{}l*{9}{Y}@{}}
\toprule

\multirow{2}{*}{\textbf{Methods}}
& \multicolumn{3}{c}{\textbf{Generated Caption}}
& \multicolumn{3}{c}{\textbf{GT Caption}}
& \multicolumn{3}{c}{\textbf{Direct Prompt}} \\

\cmidrule(lr){2-4}
\cmidrule(lr){5-7}
\cmidrule(lr){8-10}

&
$\boldsymbol{\mathcal{J}}$
&
$\boldsymbol{\mathcal{F}}$
&
$\boldsymbol{\mathcal{J}\&\mathcal{F}}$
&
$\boldsymbol{\mathcal{J}}$
&
$\boldsymbol{\mathcal{F}}$
&
$\boldsymbol{\mathcal{J}\&\mathcal{F}}$
&
$\boldsymbol{\mathcal{J}}$
&
$\boldsymbol{\mathcal{F}}$
&
$\boldsymbol{\mathcal{J}\&\mathcal{F}}$
\\

\midrule

Sa2VA-InternVL3-8B~\cite{sa2va}
& \rankthree{38.00} & \rankthree{44.04} & \rankthree{41.02}
& \rankthree{46.95} & \rankthree{53.64} & \rankthree{50.29}
& 34.68 & 40.57 & 37.63 \\

Sa2VA-InternVL3-14B~\cite{sa2va}
& \ranktwo{38.44} & \ranktwo{46.12} & \ranktwo{42.28}
& \ranktwo{49.71} & \ranktwo{57.44} & \ranktwo{53.57}
& \ranktwo{41.01} & \ranktwo{47.91} & \ranktwo{44.46} \\

UniPixel-3B~\cite{UniPixel}
& 17.81 & 21.60 & 19.70
& 39.15 & 43.07 & 41.11
& 25.72 & 29.82 & 27.77 \\

UniPixel-7B~\cite{UniPixel}
& 23.39 & 27.14 & 25.26
& 39.39 & 43.38 & 41.39
& \rankthree{37.28} & \rankthree{42.74} & \rankthree{40.01} \\

Veason-R1~\cite{veason}
& 37.33 & 42.86 & 40.10
& 46.26 & 51.62 & 48.94
& 33.68 & 38.98 & 36.33 \\

Qwen3-VL-8B-SAMTok~\cite{samtok}
& \rankone{48.65} & \rankone{55.06} & \rankone{51.86}
& \rankone{52.69} & \rankone{58.97} & \rankone{55.83}
& \rankone{43.66} & \rankone{49.39} & \rankone{46.52} \\

\bottomrule
\end{tabularx}

\caption{
Performance comparison on Anomaly Instance Reasoning Segmentation (A-IRS). 
Top three performers in each category are highlighted from 
\rankbox{topone}{Dark} (highest) to 
\rankbox{topthree}{Light} (third highest).
}

\label{tab:JF_table}

\end{table*}

%% file: Figures/Experiments/ablation_table.tex
\begin{table*}[!t]
\centering
\small
\setlength{\tabcolsep}{1mm}
\renewcommand{\arraystretch}{0.9}


\begin{tabular*}{\textwidth}{@{\extracolsep{\fill}}ccc@{}}
\begin{tabular}{lc}
\toprule
\multicolumn{2}{c}{
\textbf{L1: Instance-Level Identification}}
\\
\midrule
Variant & AI$\uparrow$\\
\midrule
Baseline
&41.32\\

AEE(Instance-Level)
&\textbf{42.56}\\

\bottomrule
\end{tabular}

&


\begin{tabular}{lcccc} 
\toprule
\multicolumn{5}{c}{
\textbf{L2: Event-Level Understanding}}
\\
\midrule
Variant
&tIoU$\uparrow$
&TI$\uparrow$
&TP$\uparrow$
&Loc$\uparrow$
\\
\midrule

Baseline
&62.45
&34.53
&33.63
&57.68
\\

AEE(Event-Level)
&\textbf{72.28}
&\textbf{38.49}
&\textbf{37.45}
&\textbf{63.72}
\\

\bottomrule
\end{tabular}

&


\begin{tabular}{lccc}
\toprule
\multicolumn{4}{c}{
\textbf{L3: Scene-Level Reasoning}}
\\
\midrule

Variant
&SF$\uparrow$
&CV$\uparrow$
&RR$\uparrow$
\\
\midrule

Baseline
&58.55
&40.52
&47.29
\\

AEE(Scene-Level)
&\textbf{60.65}
&\textbf{41.92}
&\textbf{48.94}
\\

\bottomrule
\end{tabular}

\end{tabular*}


\caption{
Ablation study of the proposed AEE strategies for hierarchical anomaly understanding based on Qwen3.5-9B.
}
\label{tab:ablation}

\end{table*}

%% file: Figures/Experiments/case_study.tex
\begin{figure*}[!t]
\centering
\includegraphics[width=0.98\textwidth]{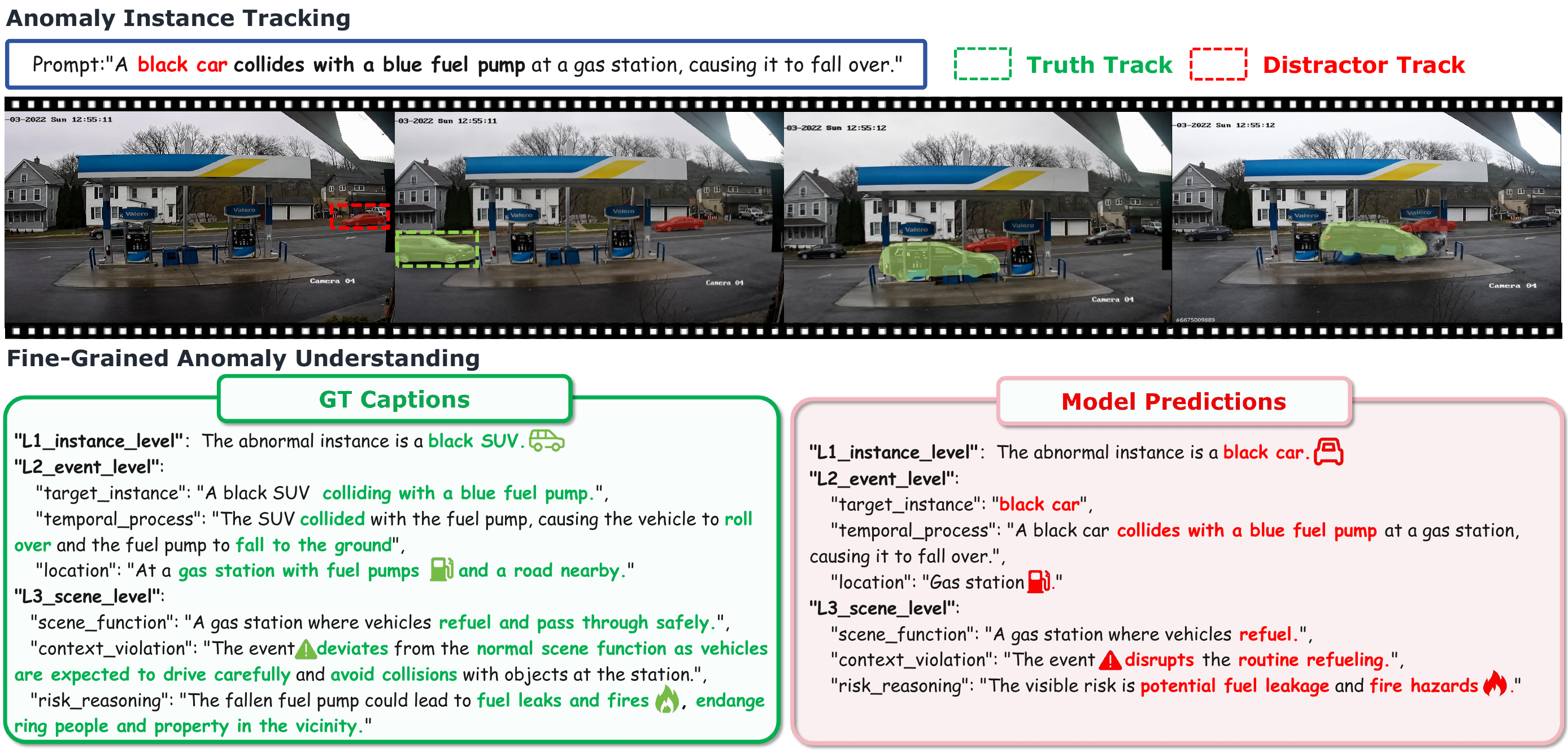}
\caption{
\textbf{A representative failure case of Sa2VA-InternVL3-14B on TAU-Bench.}
The model generates plausible event and scene-level descriptions, yet tracks a distractor vehicle rather than the ground-truth anomaly instance. This case reveals that linguistic plausibility does not necessarily imply visually grounded anomaly understanding.
}
\label{fig:case_study}
\end{figure*}

%% file: supplementary_arxiv_content.tex

\clearpage

\setcounter{section}{0}
\setcounter{subsection}{0}
\setcounter{subsubsection}{0}
\setcounter{figure}{0}
\setcounter{table}{0}
\setcounter{equation}{0}

\renewcommand{\thesection}{S\arabic{section}}
\renewcommand{\thesubsection}{\thesection.\arabic{subsection}}
\renewcommand{\thesubsubsection}{\thesubsection.\arabic{subsubsection}}
\renewcommand{\thefigure}{S\arabic{figure}}
\renewcommand{\thetable}{S\arabic{table}}
\renewcommand{\theequation}{S\arabic{equation}}

\makeatletter
\@ifpackageloaded{hyperref}{%
  \renewcommand{\theHsection}{supp.\arabic{section}}%
  \renewcommand{\theHsubsection}{supp.\arabic{section}.\arabic{subsection}}%
  \renewcommand{\theHsubsubsection}{supp.\arabic{section}.\arabic{subsection}.\arabic{subsubsection}}%
  \renewcommand{\theHfigure}{supp.\arabic{figure}}%
  \renewcommand{\theHtable}{supp.\arabic{table}}%
  \renewcommand{\theHequation}{supp.\arabic{equation}}%
}{}
\makeatother

\twocolumn[
\begin{center}
{\LARGE\bfseries
TAU-Bench: From Anomaly Instance Tracking to Fine-Grained Video Anomaly Understanding
\par}

\vspace{0.45em}

{\Large\bfseries Supplementary Materials\par}

\vspace{1.1em}
\end{center}
]

This is the supplementary material for the paper ``\textit{TAU-Bench: From Anomaly Instance Tracking to Fine-Grained Video Anomaly Understanding}.'' It provides additional details on dataset construction, quality control, evaluation protocols, metrics, and prompts. The contents are organized as follows:

\begin{itemize}[leftmargin=1.25em,itemsep=0.1em,topsep=0.2em]
\item \textbf{Section~\ref{sec:supp_dataset_construction}} describes data collection, filtering, diversity, and candidate ranking.
\item \textbf{Section~\ref{sec:supp_track_construction}} explains caption-guided proposal and event-aware anomaly instance track construction.
\item \textbf{Section~\ref{sec:supp_hierarchical_annotation}} presents the hierarchical caption annotation schema and implementation.
\item \textbf{Section~\ref{sec:supp_human_quality_control}} summarizes human review, correction, and disagreement resolution.
\item \textbf{Section~\ref{sec:supp_eval_framework}} defines the evaluation setup for detection, localization, and \airs{}.
\item \textbf{Section~\ref{sec:supp_metric_details}} details the metrics, including tracking metrics and understanding metric: \tscore{}.
\end{itemize}

\section{Dataset Construction Details}
\label{sec:supp_dataset_construction}

\subsection{Candidate Data Sources}

We collect candidate anomaly videos from both static-camera and moving-camera sources. Table~\ref{tab:data_sources} summarizes the filtering process from raw candidates to the final human-verified set.

\begin{table}[!ht]
\centering
\small
\setlength{\tabcolsep}{4pt}
\renewcommand{\arraystretch}{0.85}

\begin{tabular*}{\columnwidth}{@{\extracolsep{\fill}} l l r r r @{}}

\toprule

Source & Camera & Original & Filtered & Selected \\

\midrule
ShanghaiTech & Static & 107 & 43 & 31 \\
UBnormal & Static & 543 & 231 & 146 \\
CueBench & Moving & 3056 & 2037 & 606 \\
MSAD & Moving & 240 & 228 & 164 \\
NWPU Campus & Moving & 242 & 49 & 12 \\
TAD & Moving & 260 & 204 & 159 \\

\midrule
Total & Mixed & \textbf{4,448} & \textbf{2,792} & \textbf{1,118} \\
\bottomrule
\end{tabular*}
\caption{
Candidate data sources and filtering statistics.
Original denotes raw candidate clips collected from existing datasets;
Filtered denotes clips retained after Random Forest-based anomaly suitability filtering;
Selected denotes clips retained after human verification.
}

\label{tab:data_sources}

\end{table}

\subsection{Dataset Diversity}

\begin{figure*}[p]
    \centering
    \includegraphics[width=0.98\textwidth]{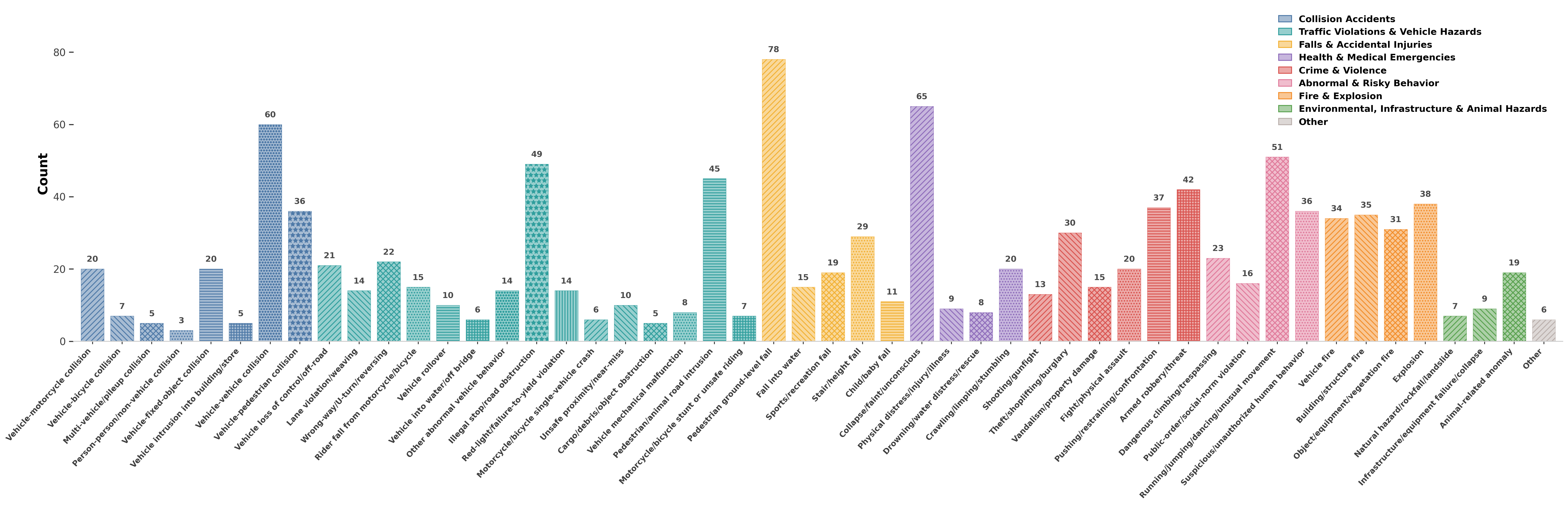}
    \vspace{1mm}
    \includegraphics[width=0.98\textwidth]{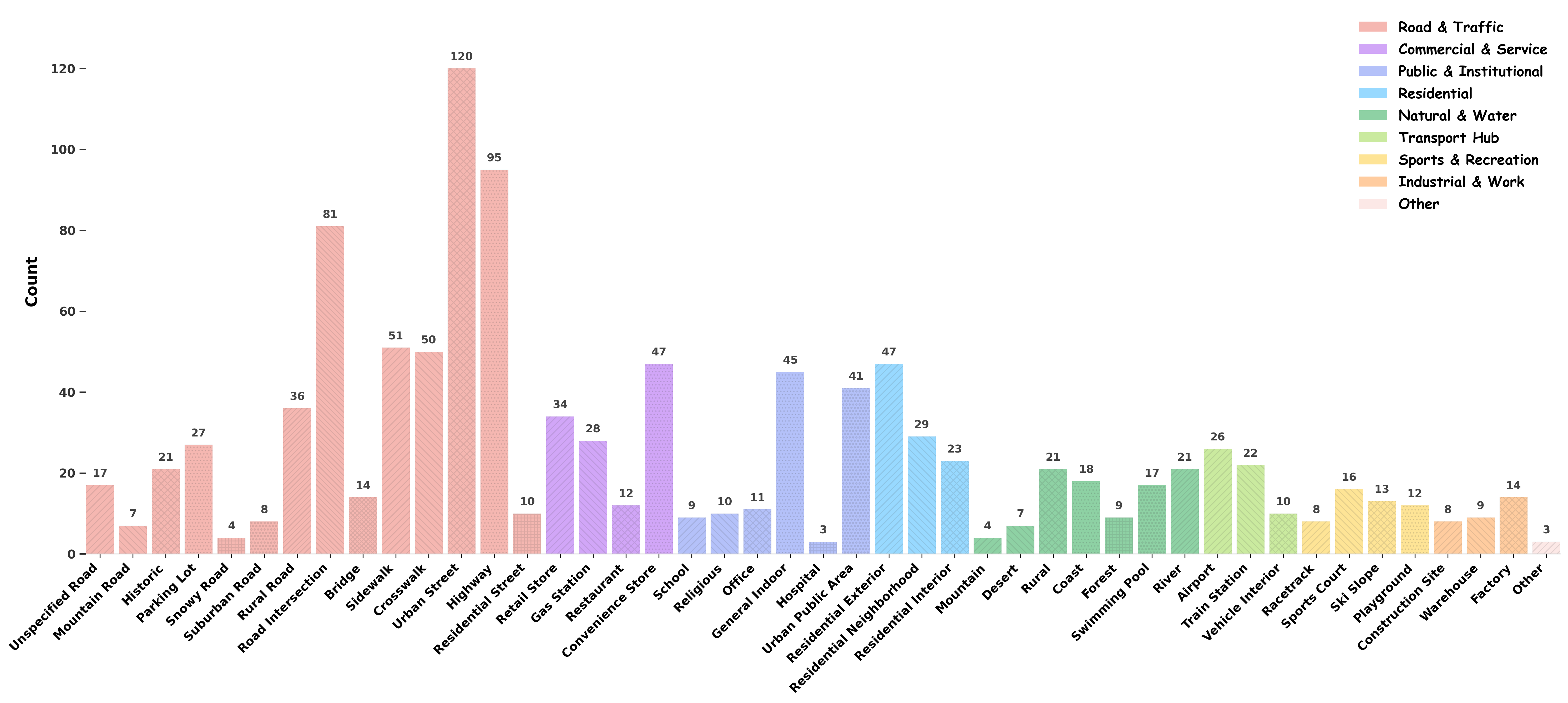}
    \caption{Event and scene category distributions in \taubench{} after human verification. The benchmark covers 49 anomaly event categories and 45 scene categories.}
    \label{fig:category_distribution}
\end{figure*}

\taubench{} covers diverse anomaly types and scene contexts rather than a single surveillance setting. Fig.~\ref{fig:category_distribution} reports the human-verified event and scene distributions, including traffic accidents, unsafe behaviors, falls, medical emergencies, fire-related events, violence, environmental hazards, and road, residential, public, natural, commercial, and recreational scenes.

\subsection{Anomaly Suitability Assessment}

The automatic filtering stage scores annotation suitability rather than abnormality alone. We label 500 calibration videos with human suitability scores in $\{0,1,2\}$, extract VLM QA scores in $[0,5]$ over six evidence dimensions, and train a Random Forest regressor to rank the remaining candidates.

\subsubsection{Human Labels.}
The 500 calibration videos are labeled as:

\begin{itemize}
  \item \textbf{0: Normal or unsuitable.} The video contains no identifiable anomaly evidence or lacks sufficient evidence for instance-level annotation.
  \item \textbf{1: Ambiguous or difficult.} The video contains potential anomaly evidence, but the focal instance, temporal boundary, contextual interpretation, or risk source remains ambiguous.
  \item \textbf{2: Clear anomaly candidate.} The video contains clear anomaly evidence with an identifiable focal instance, trackable visual cues, and interpretable contextual or risk information.
\end{itemize}

\subsubsection{VLM QA Assessment.}
Each VLM QA feature is scored from 0 to 5, where 0 denotes unusable evidence and 5 denotes clear evidence. Raw question-level scores are used as Random Forest inputs; dimension averages are used only for analysis.


\begin{table*}[p]
\centering
\small
\setlength{\tabcolsep}{4pt}
\renewcommand{\arraystretch}{1.15}

\begin{tabularx}{\textwidth}{L{0.18\textwidth} X X}

\toprule
Dimension & QA Features & Description\\
\midrule

Visual Quality &
Visual clarity
&
Measures whether the video quality supports reliable visual annotation.
\\

Instance Visibility &
Subject clarity; annotation suitability; temporal visibility; instance identifiability
&
Measures whether the target instance can be localized, segmented, and tracked over time.
\\

Scene Context &
Scene function; spatial constraints; normal-reference contrast
&
Measures whether scene-level functional and spatial context supports anomaly interpretation.
\\

Instance-Scene Incongruity &
Area mismatch; behavior-scene conflict; contextual inappropriateness; non-category evidence; grounded explanation; normal plausibility ambiguity
&
Measures whether anomaly judgments are grounded in instance-level inconsistency rather than category priors.
\\

Risk Evidence &
Risk proximity; collision evidence; aggressive interaction; flow disruption; hazard events; unsafe posture; static-scene interaction
&
Measures visible evidence of hazards, unsafe behaviors, abnormal interactions, or scene disruption.
\\

Risk Attribution &
Source reaction; group reaction; source-target clarity
&
Measures whether the risk source and affected entities can be identified.
\\

\bottomrule

\end{tabularx}

\caption{
Six VLM-assessed dimensions for anomaly suitability assessment.
}
\label{tab:qa_dimensions}

\end{table*}

\begin{figure*}[p]
    \centering
    \includegraphics[width=0.9\textwidth]{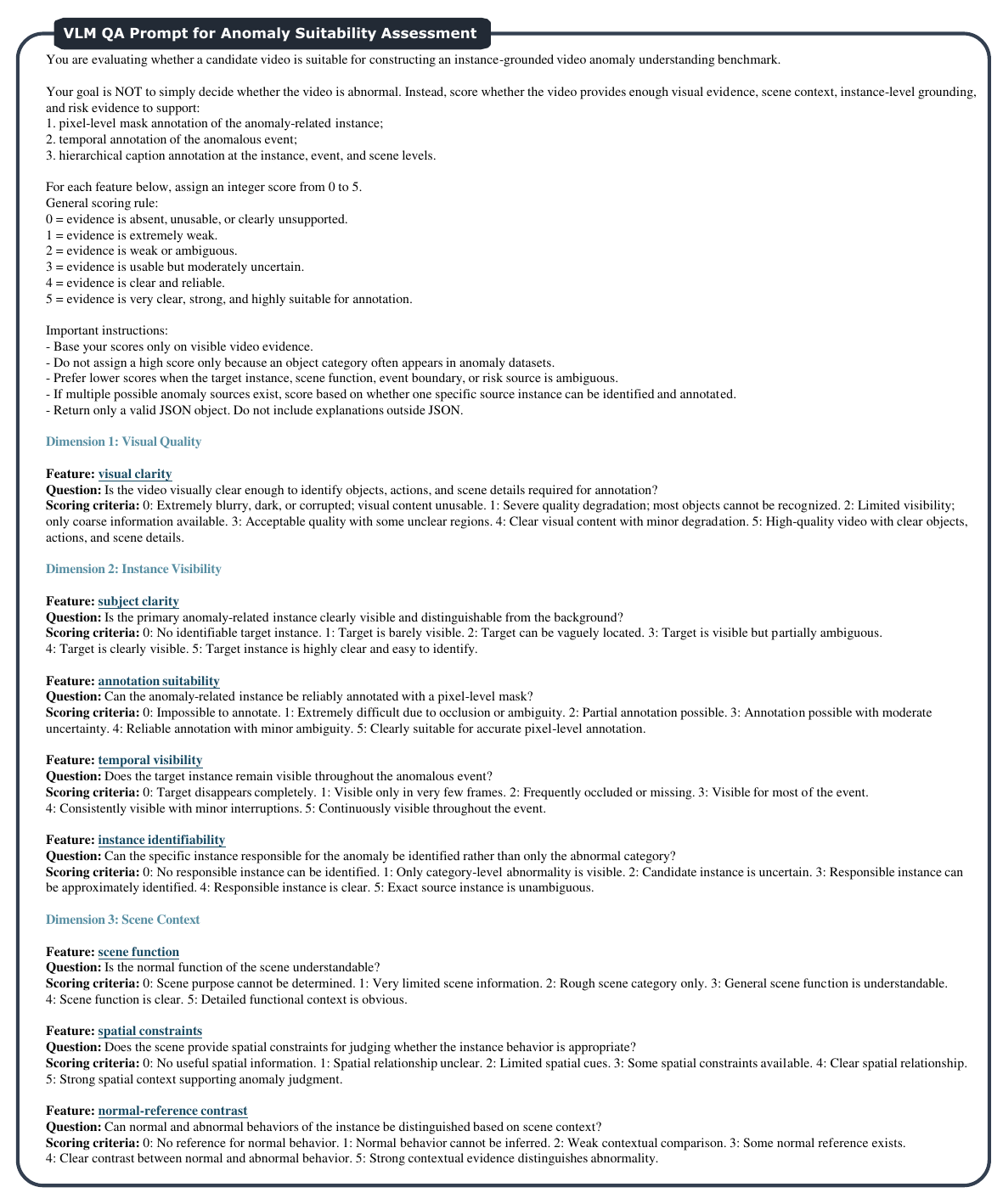}
    \caption{VLM QA prompt for anomaly suitability assessment (Part I), covering visual quality, instance visibility, and scene context.}
    \label{fig:qa_prompt_part1}
\end{figure*}

\begin{figure*}[p]
    \centering
    \includegraphics[width=0.9\textwidth]{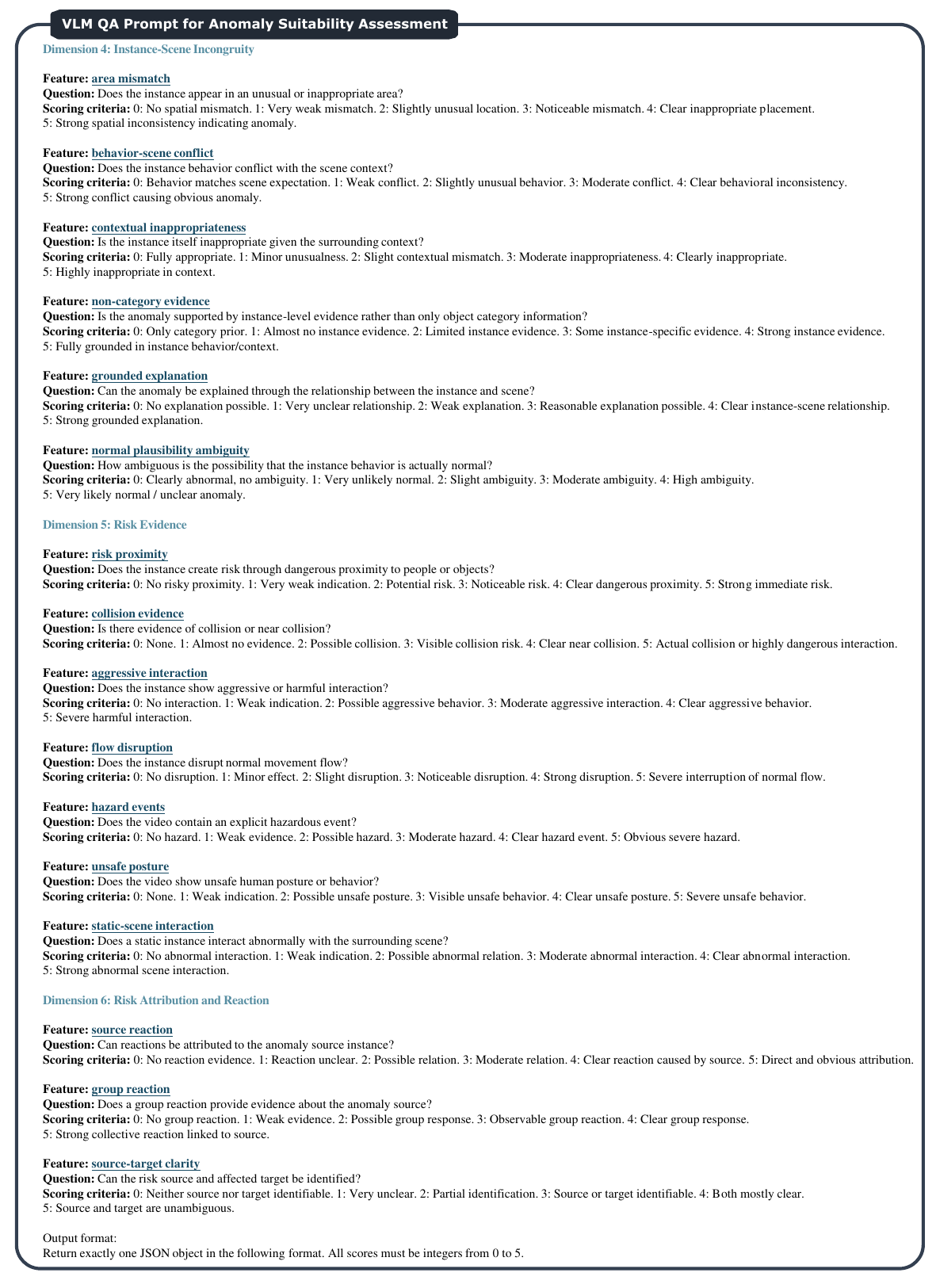}
    \caption{VLM QA prompt for anomaly suitability assessment (Part II), covering instance-scene incongruity, risk evidence, and risk attribution.}
    \label{fig:qa_prompt_part2}
\end{figure*}

The structured VLM prompt asks the model to score visual evidence, contextual information, and risk cues for instance-grounded annotation. The full JSON prompt is provided in 
Figs.~\ref{fig:qa_prompt_part1} and \ref{fig:qa_prompt_part2},
covering all six suitability assessment dimensions.




\subsubsection{Random Forest Details.}

The Random Forest regressor uses VLM QA scores as features and human suitability scores as ordered regression targets. We use 300 trees, maximum depth 12, random seed 42, all CPU workers, and an 80/20 stratified split of the 500 labeled videos. Candidates with predicted suitability score $\geq1.5$ are retained for track construction and human verification.

\section{Anomaly Instance Track Construction}
\label{sec:supp_track_construction}

\subsection{Caption-Guided Track Proposal}

For each retained clip, a pretrained vision-language model Vad-R1 first generates a coarse who-what caption. The \textit{who} part describes the visible anomaly instance, and the \textit{what} part summarizes its anomalous action. The \textit{who} description is then used as a text prompt for SAM3 to generate temporally linked candidate mask tracks.

For each candidate track $\tau$, we sample masked crops that isolate the target instance and context crops that preserve the surrounding event information. The masked crops are used for appearance matching, while context crops are used for event matching.

\subsection{Event-Aware Track Selection for Mask-Track Construction}
\label{sec:event-aware-track-selection}

SAM3 often produces multiple candidate tracks for a single anomaly prompt. We therefore use a training-free Event-Aware Track Selection module to generate initial mask-track proposals, which are then verified by human annotators. Given the instance description $q_{\mathrm{who}}$, the event description $q_{\mathrm{what}}$, and a set of candidate tracks $\mathcal{C}$, the module scores each candidate track $\tau\in\mathcal{C}$ from four aspects: appearance consistency, event consistency, motion evidence, and track quality.

For each track, we uniformly sample at most $M=12$ frames. In each sampled frame $t$, we extract a masked crop $I_t^{m}$ using the candidate mask and a context crop $I_t^{c}$ by expanding the mask bounding box by a factor of 1.8. We use CLIP image-text similarity
\begin{equation}
\phi(I,q)=
\mathrm{clip}\left(
\frac{\cos(\mathrm{CLIP}_{I}(I),\mathrm{CLIP}_{T}(q))+1}{2},
0,1
\right).
\end{equation}
Let $\mathrm{TopKAvg}_{5}(\cdot)$ denote the average of the top-5 frame scores. The appearance score is
\begin{equation}
S_{\mathrm{app}}(\tau)
=
\mathrm{TopKAvg}_{5}
\left(
\{\phi(I_t^{m}, q_{\mathrm{who}})\}_{t\in\tau}
\right),
\end{equation}
and the event score is
\begin{equation}
S_{\mathrm{evt}}(\tau)
=
\mathrm{TopKAvg}_{5}
\left(
\{\phi(I_t^{c}, q_{\mathrm{evt}})\}_{t\in\tau}
\right),
\end{equation}
where $q_{\mathrm{evt}}$ concatenates the target-instance description and the event description.

For motion scoring, let $b_t=(x^1_t,y^1_t,x^2_t,y^2_t)$ be the bounding box of the candidate mask, $c_t$ be its center, $a_t$ be its area, $r_t$ be its aspect ratio, and $D=\sqrt{W^2+H^2}$ be the image diagonal. We compute

\begin{align}
d_{\mathrm{disp}} &= \mathrm{clip}\left(4\frac{\|c_T-c_1\|_2}{D},0,1\right),\\
d_{\mathrm{path}} &= \mathrm{clip}\left(2\frac{\sum_{t=2}^{T}\|c_t-c_{t-1}\|_2}{D},0,1\right),\\
d_{\mathrm{area}} &= \mathrm{clip}\left(\frac{\max_t a_t-\min_t a_t}{\mathrm{mean}_t(a_t)},0,1\right),\\
d_{\mathrm{ratio}} &= \mathrm{clip}\left(\frac{\max_t r_t-\min_t r_t}{\mathrm{mean}_t(r_t)},0,1\right).
\end{align}
The motion score is
\begin{equation}
S_{\mathrm{mot}}(\tau)
=
0.35d_{\mathrm{disp}}
+
0.35d_{\mathrm{path}}
+
0.15d_{\mathrm{area}}
+
0.15d_{\mathrm{ratio}}.
\end{equation}

The quality score measures whether the candidate track has sufficient valid masks, temporal continuity, and reasonable mask size. Let $n_v$ be the number of valid sampled masks, $n_s=\min(M,|\tau|)$ be the number of sampled frames, and $\mathcal{F}_{\tau}$ be the set of frame indices covered by the track. We define
\begin{equation}
q_{\mathrm{valid}}=\frac{n_v}{n_s},
\end{equation}
\begin{equation}
q_{\mathrm{cont}}
=
\mathrm{clip}
\left(
5\frac{|\mathcal{F}_{\tau}|}
{\max(\mathcal{F}_{\tau})-\min(\mathcal{F}_{\tau})+1},
0,1
\right).
\end{equation}
For mask area, let $\bar{a}$ be the average mask area and $A=WH$ be the image area:
\begin{equation}
q_{\mathrm{area}}
=
\begin{cases}
0.1, & \bar{a}/A < 0.0002,\\
0.2, & \bar{a}/A > 0.7,\\
1.0, & \mathrm{otherwise}.
\end{cases}
\end{equation}
The final quality score is
\begin{equation}
S_{\mathrm{qual}}(\tau)
=
0.45q_{\mathrm{valid}}
+
0.35q_{\mathrm{cont}}
+
0.20q_{\mathrm{area}}.
\end{equation}

The final track score is
\begin{equation}
S(\tau)
=
0.35S_{\mathrm{app}}(\tau)
+
0.40S_{\mathrm{evt}}(\tau)
+
0.15S_{\mathrm{mot}}(\tau)
+
0.10S_{\mathrm{qual}}(\tau).
\end{equation}
We select the highest-scoring candidate:
\begin{equation}
\hat{\tau}
=
\arg\max_{\tau\in\mathcal{C}} S(\tau).
\end{equation}
Let $S_{(1)}$ and $S_{(2)}$ be the highest and second-highest scores. The confidence margin is
\begin{equation}
m=S_{(1)}-S_{(2)}.
\end{equation}
A proposal is marked as low-confidence if $S_{(1)}<0.45$ and ambiguous if $m<0.08$.

All automatically selected tracks are manually verified. Across 1,118 videos, we obtain 1,454 final anomaly-instance tracks. Among them, 880 tracks are selected by the automatic module and accepted after human verification, while the remaining 574 tracks are re-annotated or refined frame by frame using XLabelAnything. The automatic proposal acceptance rate is therefore
\begin{equation}
\mathrm{AcceptRate}
=
\frac{880}{1454}
=
60.5\%,
\end{equation}
and the manual frame-level annotation rate is
\begin{equation}
\mathrm{ManualRate}
=
\frac{574}{1454}
=
39.5\%.
\end{equation}
Thus, 60.5\% of final tracks are accepted from automatic proposals, while the remaining 39.5\% are completed by human frame-level repair.

\section{Hierarchical Caption Annotation}
\label{sec:supp_hierarchical_annotation}

\subsection{Annotation Schema}

Each video $V$ is annotated with hierarchical captions:
\begin{equation}
H(V) = \{c_{\mathrm{ins}}, c_{\mathrm{evt}}, c_{\mathrm{scn}}\}.
\end{equation}
The three levels are independently supervised but semantically progressive. All fields must refer to the same anomaly instance.

\begin{table}[t]
\centering
\small
\begin{tabularx}{\linewidth}{l l X}
\toprule
Level & Field & Description \\
\midrule
Instance-Level & anomalous\_instance & Visual category and appearance of the focal anomaly instance. \\
Event-Level & target\_instance & The same focal instance described in relation to the anomalous event. \\
Event-Level & temporal\_process & The temporal evolution of the anomalous event. \\
Event-Level & location & The physical location or local scene where the event occurs. \\
Scene-Level & scene\_function & The normal expected function of the scene. \\
Scene-Level & context\_violation & How the event violates the expected scene function or contextual norm. \\
Scene-Level & risk\_reasoning & Potential danger, injury, or disruption caused by the event. \\
\bottomrule
\end{tabularx}
\caption{Hierarchical annotation schema used in \taubench.}
\label{tab:caption_schema}
\end{table}

\subsection{Implementation Details}
\label{sec:supp_hierarchical_caption}

TAU-Bench constructs three progressive annotations,
\begin{equation}
\mathcal{H}(V)
=
\left\{
c_{\mathrm{ins}},
c_{\mathrm{evt}},
c_{\mathrm{scn}}
\right\},
\end{equation}
where $c_{\mathrm{ins}}$ identifies the focal anomaly instance,
$c_{\mathrm{evt}}$ describes its complete visible event, and
$c_{\mathrm{scn}}$ interprets the event under scene-specific expectations.
All levels share the same focal visual referent and are verified against the corrected anomaly track.

\begin{figure*}[t]
    \centering
    \includegraphics[width=1\textwidth]{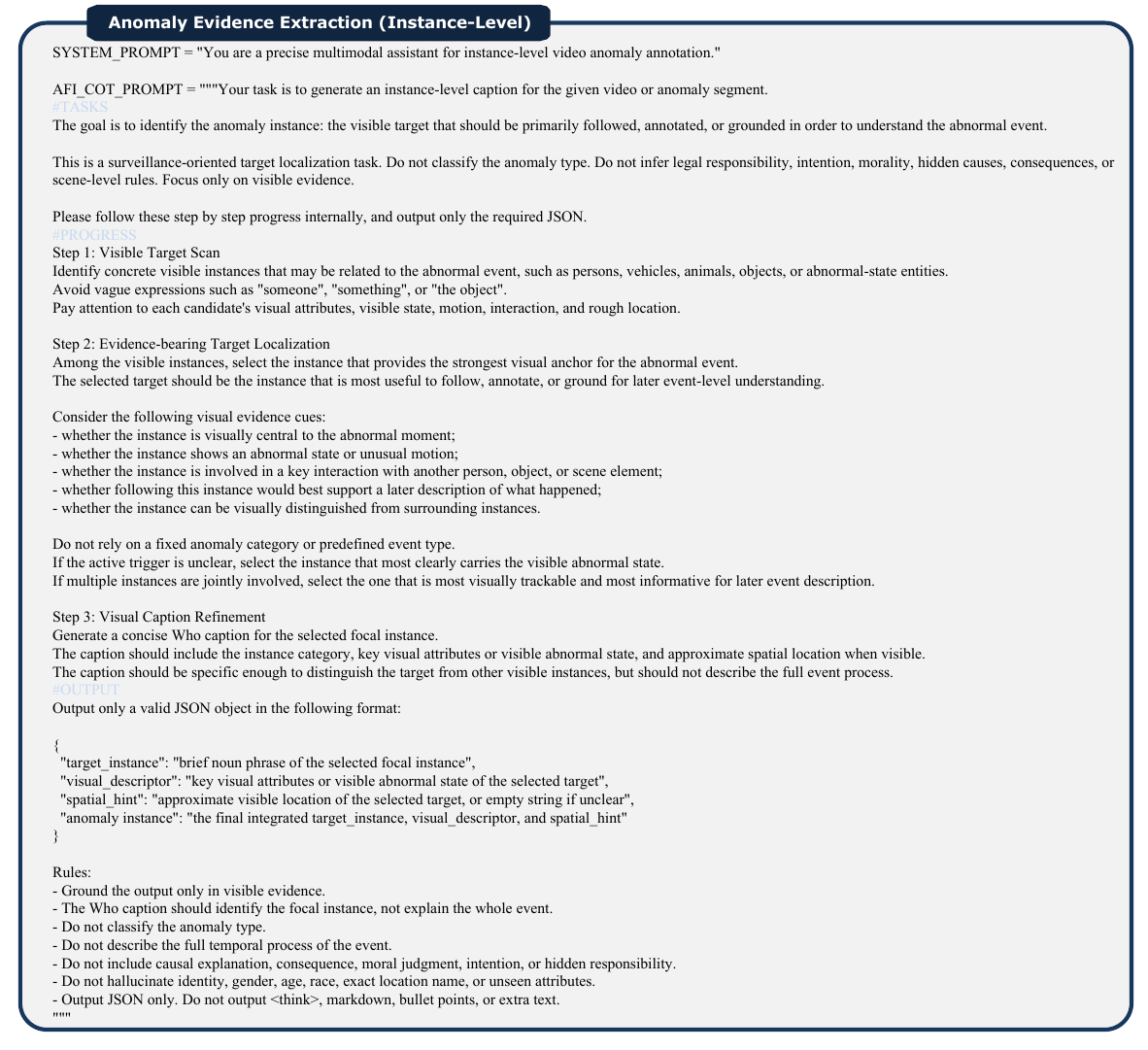}
   \caption{Prompt template for instance-level identification, which guides the VLM to select the focal anomaly instance from visible candidates and describe its category, appearance, state, and spatial cues.}
    \label{fig:l1_prompt}
\end{figure*}

\paragraph{L1: Instance-Level Identification.}
The full video is provided to the VLM to identify the most distinctive and trackable anomaly instance. The output is a concise instance caption, without event narration or scene-level reasoning. The prompt is shown in Fig.~\ref{fig:l1_prompt}.

\paragraph{L2: Event-Level Understanding.}
We first localize the anomalous interval and then generate an event caption. Given video duration $D$, we divide the video into overlapping windows,
\begin{equation}
w_i =
\left[
i\Delta,
\min(i\Delta+L,D)
\right],
\end{equation}
with $L=5$ seconds and $\Delta=4$ seconds. We sample 16 frames per window and query Qwen3.6-35B for a normal/abnormal decision, anomaly score $s_i\in[0,1]$, visible subject, event observation, location, and visual cue.

Scores are smoothed by a median filter with kernel size three:
\begin{equation}
m_i =
\operatorname{median}
\left(
s_{i-1},s_i,s_{i+1}
\right).
\end{equation}
and an exponential moving average:
\begin{equation}
\bar{s}_i =
\alpha m_i +
(1-\alpha)\bar{s}_{i-1},
\qquad
\alpha=0.4.
\end{equation}

Threshold aggregation localizes event intervals. An interval starts when
$\bar{s}_i\geq\theta_{\mathrm{high}}$, where
$\theta_{\mathrm{high}}=0.65$, and continues through adjacent windows until
the score falls below $\theta_{\mathrm{low}}=0.40$. Intervals shorter than one second are discarded, and intervals separated by at most two seconds are merged.

\begin{figure*}[t]
    \centering
    \includegraphics[width=1\textwidth]{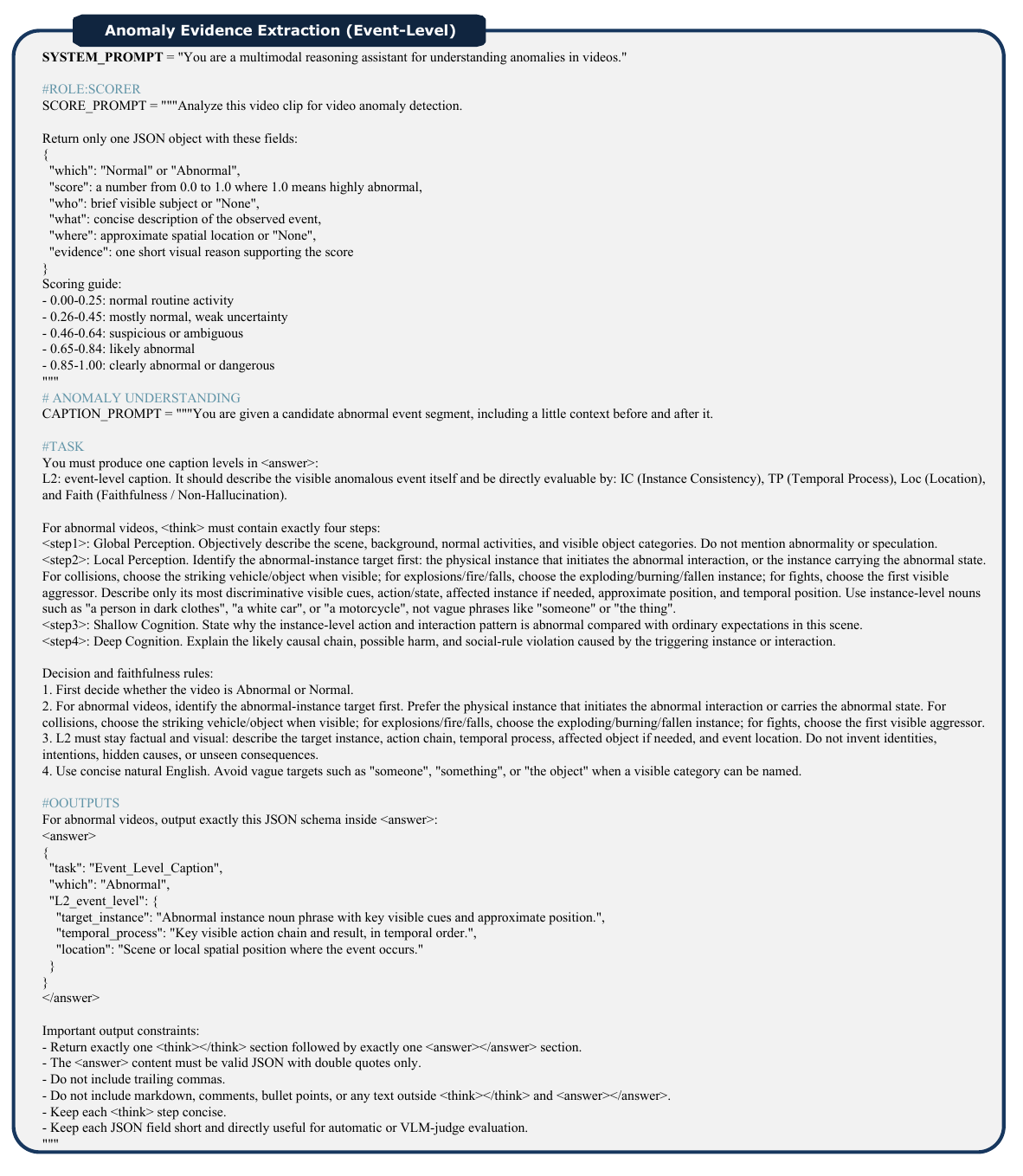}
    \caption{Prompt templates for window anomaly scoring and event-level caption generation, covering temporal localization and structured descriptions of the target instance, temporal process, and location.}
    \label{fig:l2_prompt}
\end{figure*}

Each localized segment is extended with two seconds of temporal context on
both sides:
\begin{equation}
e_j^{+}
=
\left[
\max(0,t_j^s-2),
\min(D,t_j^e+2)
\right].
\end{equation}
We sample 32 frames from $e_j^{+}$ for structured event captioning. Generation uses temperature $0.1$, top-$p$ $0.9$, and at most 512 output tokens. The prompts are shown in Fig.~\ref{fig:l2_prompt}.

\paragraph{L3: Scene-Level Reasoning.}
L3 explains why the localized event violates the scene's normal function. To preserve identity consistency, event support is derived from the verified anomaly track. Let $\Omega^{*}$ be the valid-mask frame indices of the verified track; its span is
\begin{equation}
t_s^{*}=\min\Omega^{*},
\qquad
t_e^{*}=\max\Omega^{*}.
\end{equation}
Four event-anchor frames are uniformly selected from $\Omega^{*}$. Frames outside the track-supported span form the context candidate set:
\begin{equation}
\mathcal{C}
=
\left\{
I_t
\mid
t<t_s^{*}
\ \lor\
t>t_e^{*}
\right\}.
\end{equation}
Pre- and post-event frames are sampled every four frames and supplemented with boundary frames.

Each context frame is encoded by Qwen3-VL-8B-Instruct. We mean-pool and $\ell_2$-normalize the visual tokens:
\begin{equation}
\mathbf{m}_t
=
\operatorname{Norm}
\left(
\operatorname{MeanPool}
\left(
E_v(I_t)
\right)
\right),
\end{equation}
where $\mathbf{m}_t$ is a context memory unit. The bank is limited to 48 visually diverse frames selected by cosine distance.

The structured L2 fields are then mapped to three evidence queries:
\begin{align}
y_{\mathrm{loc}}
&\rightarrow q_{\mathrm{func}},
\\
y_{\mathrm{target}}
&\rightarrow q_{\mathrm{rule}},
\\
y_{\mathrm{process}}
&\rightarrow q_{\mathrm{risk}}.
\end{align}
The three queries retrieve scene function, normal behavioral rule, and risk-consequence evidence.

For each query $q_d$, we mean-pool and normalize its token embeddings to obtain
$\mathbf{q}_d$. A context memory unit is ranked as
\begin{equation}
r_d(t)
=
\cos
\left(
\mathbf{q}_d,\mathbf{m}_t
\right)
+
\beta_d(\rho_t),
\end{equation}
where $\rho_t$ denotes pre-/post-event role and $\beta_d$ is a temporal prior. The pre/post priors are $(0.40,0.05)$ for scene function, $(0.50,0.00)$ for normal rule, and $(0.00,0.50)$ for risk consequence. We select the top four frames per query.

Retrieved frames are deduplicated and combined with four event-anchor frames under a 24-frame budget. The prompt includes L2 fields, track identity, and retrieved frame indices. Qwen3.6-35B then identifies scene function, expected rule, context violation, and visible or cautiously inferred risk. Generation uses temperature $0.1$, top-$p$ $0.9$, and at most 768 output tokens.

The resulting scene-level annotation is
\begin{equation}
c_{\mathrm{scn}}
=
\left\{
y_{\mathrm{cap}},
y_{\mathrm{func}},
y_{\mathrm{vio}},
y_{\mathrm{risk}}
\right\},
\end{equation}
corresponding to \texttt{caption}, \texttt{scene\_function}, \texttt{context\_dependency}, and \texttt{risk\_reasoning}. The implementation key \texttt{context\_dependency} corresponds to the Context Violation field. Confidence labels are retained for inspection but excluded from benchmark scoring.

\paragraph{Cross-Level Verification.}
The generated hierarchy is reviewed with the raw video, verified interval, and corrected track. Annotators check identity consistency across L1--L3, event completeness, scene-context support, and risk plausibility. Unsupported statements are removed; unresolved identity ambiguity or unsupported scene reasoning leads to rejection.

\section{Human Quality Control}
\label{sec:supp_human_quality_control}


\begin{table*}[!t]
\centering
\small

\setlength{\tabcolsep}{4pt}
\renewcommand{\arraystretch}{1.12}

\begin{tabular}{p{0.15\textwidth}
                p{0.13\textwidth}
                @{\hspace{1pt}}
                p{0.50\textwidth}
                p{0.16\textwidth}}
\toprule
Stage & Scope & Annotator Allocation and Review Protocol & Output \\
\midrule
Video verification
& 2,792 videos
& Three trained annotators screened candidate videos; one of them adjudicated the final retained set.
& Candidated videos. \\

Interval verification
& 1,219 intervals
& Two trained annotators reviewed temporal boundaries; minor boundary differences were averaged.
& Verified anomaly intervals. \\

Track verification
& 1,454 tracks
& Three annotators inspected identity consistency, temporal completeness, and mask validity; uncertain cases were sent to a reviewer.
& 880 verified tracks. \\

Mask annotation
& 574 tracks
& Two annotators completed tracks that could not be reliably obtained automatically; two reviewers checked corrected masks.
& Completed instance-level masks. \\

Caption verification
& 460 videos
& Three annotators revised fine-grained captions and checked semantic consistency across instance-, event-, and scene-level fields.
& Final corrected captions. \\

\midrule
\textbf{Total}
& --
& Five trained annotators participated in the overall review pipeline.
& Final annotations. \\
\bottomrule
\end{tabular}
\caption{Human review workflow. Automatically generated annotations are treated as proposals and are verified, corrected, or discarded through a multi-stage quality-control process.}
\label{tab:human_review_workflow}

\vspace{15pt}


\setlength{\tabcolsep}{5pt}
\renewcommand{\arraystretch}{1.15}
\begin{tabular}{lrrrrrrr}
\toprule
Branch &
Proposals &
\makecell{Accepted\\w/o Repair} &
\makecell{Manually\\Repaired} &
\makecell{Engine\\Runtime} &
\makecell{Fully Manual\\Labor} &
\makecell{Actual Repair\\Labor} &
\makecell{Labor Saved} \\
\midrule
Caption
& 1,118
& 658 (58.9\%)
& 460 (41.1\%)
& 11.0 h
& 65.2 h
& 26.8 h
& 38.4 h (58.9\%) \\

Mask track
& 1,454
& 880 (60.5\%)
& 574 (39.5\%)
& 14.0 h
& 132.1 h
& 52.1 h
& 80.0 h (60.5\%) \\
\midrule
\textbf{Overall}
& \textbf{2,572}
& \textbf{1,538 (59.8\%)}
& \textbf{1,034 (40.2\%)}
& \textbf{25.0 h}
& \textbf{197.3 h}
& \textbf{79.0 h}
& \textbf{118.3 h (60.0\%)} \\
\bottomrule
\end{tabular}
\caption{Efficiency of the automatic annotation engine. Approximately 60\% of the generated caption and mask-track proposals are accepted without repair, reducing the estimated manual correction workload from 197.3 to 79.0 annotator-hours. Engine runtime and human labor are reported separately.}
\label{tab:annotation_efficiency}
\end{table*}


\subsection{Annotation Efficiency}


Table~\ref{tab:annotation_efficiency} reports the effective proposal quality of the automatic data engine. Among 2,572 generated caption and mask-track proposals, 1,538 (59.8\%) are accepted without repair. Under the observed correction rates, automatic proposals reduce manual correction from 197.3 to 79.0 annotator-hours, saving 118.3 hours (60.0\%).

\subsection{Verification Criteria}

Annotators check the following criteria:

\begin{itemize}[leftmargin=*]
  \item \textbf{Anomaly visibility:} the anomaly must be visually observable.
  \item \textbf{Instance uniqueness:} the focal instance must be identifiable and separable from distractors.
  \item \textbf{Track consistency:} the mask track must preserve the same identity over time.
  \item \textbf{Event completeness:} the temporal interval must cover the full anomalous process.
  \item \textbf{Contextual interpretability:} the scene must support scene-function and context-violation reasoning.
  \item \textbf{Caption faithfulness:} every caption field must be supported by visible evidence or reasonable scene context.
\end{itemize}

\subsection{Correction Operations}

Annotators may perform the following operations:

\begin{itemize}[leftmargin=*]
  \item correct identity switches by replacing masks with the correct instance;
  \item remove masks belonging to distractor objects;
  \item add missing masks when the focal instance is visible but not segmented;
  \item adjust the start and end frame of the anomaly interval;
  \item rewrite unsupported or overly speculative caption fields;
  \item discard samples with unresolved ambiguity or irreparable tracks.
\end{itemize}

\subsection{Secondary Review and Disagreement Resolution}

Retained samples are reviewed according to the stage-specific annotator allocation in Table~\ref{tab:human_review_workflow}. Samples requiring major edits or receiving low-confidence judgments are assigned to a secondary reviewer. Disagreements are resolved according to the following priority order:

\begin{enumerate}[leftmargin=*]
  \item visible identity consistency of the anomaly instance;
  \item temporal completeness of the anomalous event;
  \item consistency between the instance track and all caption fields;
  \item removal of unsupported causal or risk statements.
\end{enumerate}

\begin{figure*}[p]
    \centering
    \includegraphics[width=1\textwidth]{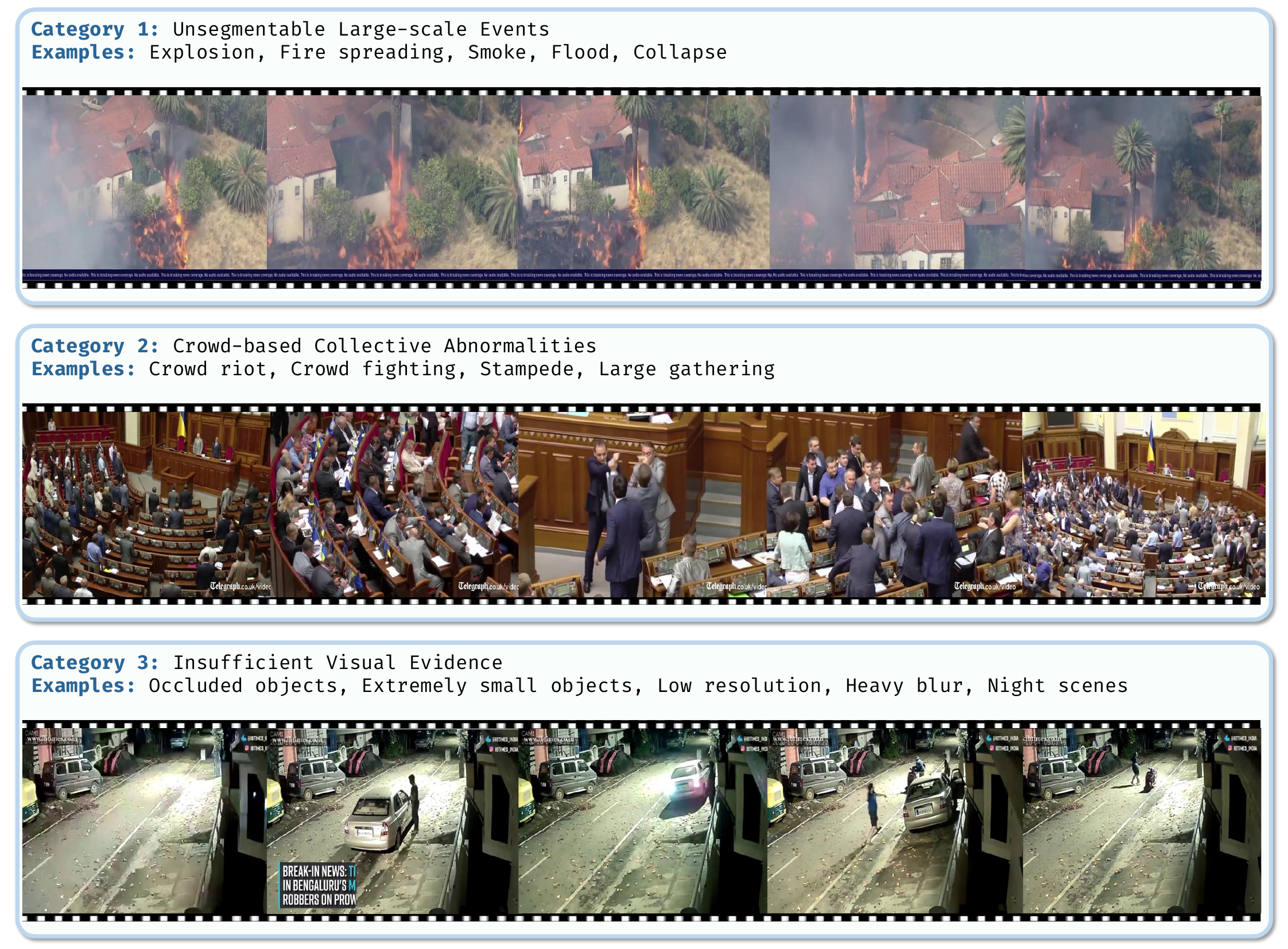}
    \caption{Representative videos excluded during manual verification. We remove samples involving unsegmentable large-scale events, crowd-based collective abnormalities without a unique focal instance, and insufficient visual evidence caused by occlusion, small targets, low resolution, blur, or poor illumination.}
    \label{fig:filter}
\end{figure*}

Figure~\ref{fig:filter} presents representative videos excluded during manual verification.

\section{Evaluation Framework}
\label{sec:supp_eval_framework}

\taubench{} evaluates anomaly instance tracking through three complementary
tasks: video-level anomaly detection, temporal anomaly localization, and
anomaly instance reasoning segmentation (\airs{}). All tasks are evaluated in a
zero-shot setting with task-specific prompts and deterministic output parsing.
Table~\ref{tab:eval_protocol_overview} summarizes the input, output, metric,
and diagnostic purpose of each protocol.

\begin{table*}[t]
\centering
\small
\setlength{\tabcolsep}{4pt}
\renewcommand{\arraystretch}{1.12}
\begin{tabularx}{\textwidth}{L{0.17\textwidth} L{0.22\textwidth} L{0.20\textwidth} L{0.13\textwidth} X}
\toprule
Task & Input & Required Output & Metric & Diagnostic Purpose \\
\midrule
Video-level anomaly detection
& Uniformly sampled keyframes from the full video
& JSON object with \texttt{which}, \texttt{confidence}, and \texttt{reason}
& Recall
& Tests whether the model detects visible abnormal evidence from sparse video observations. \\

Temporal anomaly localization
& Uniformly sampled keyframes with timestamps
& JSON list of abnormal intervals with start time, end time, and confidence
& tIoU
& Tests whether the model localizes the visible temporal extent of the anomalous event. \\

\airs{} with direct prompt
& Video frames from the annotated anomaly interval
& Mask track of the primary abnormal instance
& $\mathcal{J}$, $\mathcal{F}$, $\mathcal{J}\&\mathcal{F}$
& Tests end-to-end reasoning segmentation without an explicit semantic intermediate. \\

\airs{} with generated-caption prompt
& Video frames and the model-generated target-instance phrase
& Mask track guided by the generated instance description
& $\mathcal{J}$, $\mathcal{F}$, $\mathcal{J}\&\mathcal{F}$
& Tests whether the model's own semantic description transfers to spatial grounding. \\

\airs{} with GT-caption prompt
& Video frames and the ground-truth target-instance phrase
& Mask track guided by the reference instance description
& $\mathcal{J}$, $\mathcal{F}$, $\mathcal{J}\&\mathcal{F}$
& Isolates visual grounding ability under correct semantic guidance. \\
\bottomrule
\end{tabularx}
\caption{Evaluation protocols for anomaly instance tracking. The three \airs{} settings separate failures caused by semantic target selection from failures caused by mask-level visual grounding.}
\label{tab:eval_protocol_overview}
\end{table*}

\subsection{Video-Level Anomaly Detection}

\begin{figure*}[t]
    \centering
    \includegraphics[width=0.95\textwidth]{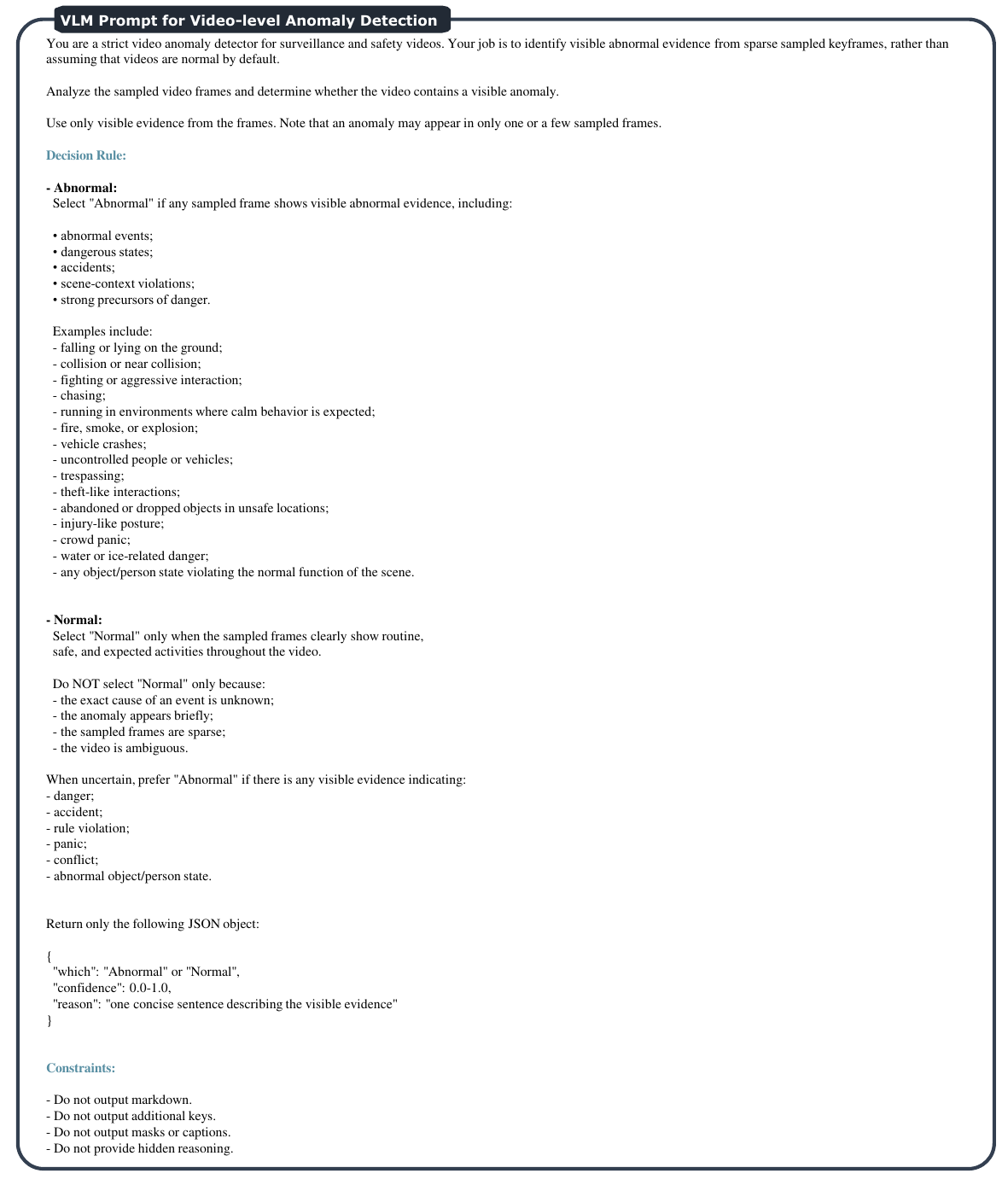}
    \caption{Prompt template for video-level anomaly detection. The model predicts whether sampled keyframes contain visible abnormal evidence and returns a structured JSON response.}
    \label{fig:vad_eval_prompt}
\end{figure*}

The model receives uniformly sampled keyframes and predicts whether visible abnormal evidence is present. The prompt emphasizes observable cues rather than hidden causes or intentions, and the full template is shown in Fig.~\ref{fig:vad_eval_prompt}.

The required output is a JSON object:
\begin{equation}
\{\texttt{which},\texttt{confidence},\texttt{reason}\}.
\end{equation}
Only \texttt{which} is used for scoring. Since all samples contain annotated anomaly evidence, performance is reported as anomaly recall.

\subsection{Temporal Anomaly Localization}

\begin{figure*}[t]
    \centering
    \includegraphics[width=0.95\textwidth]{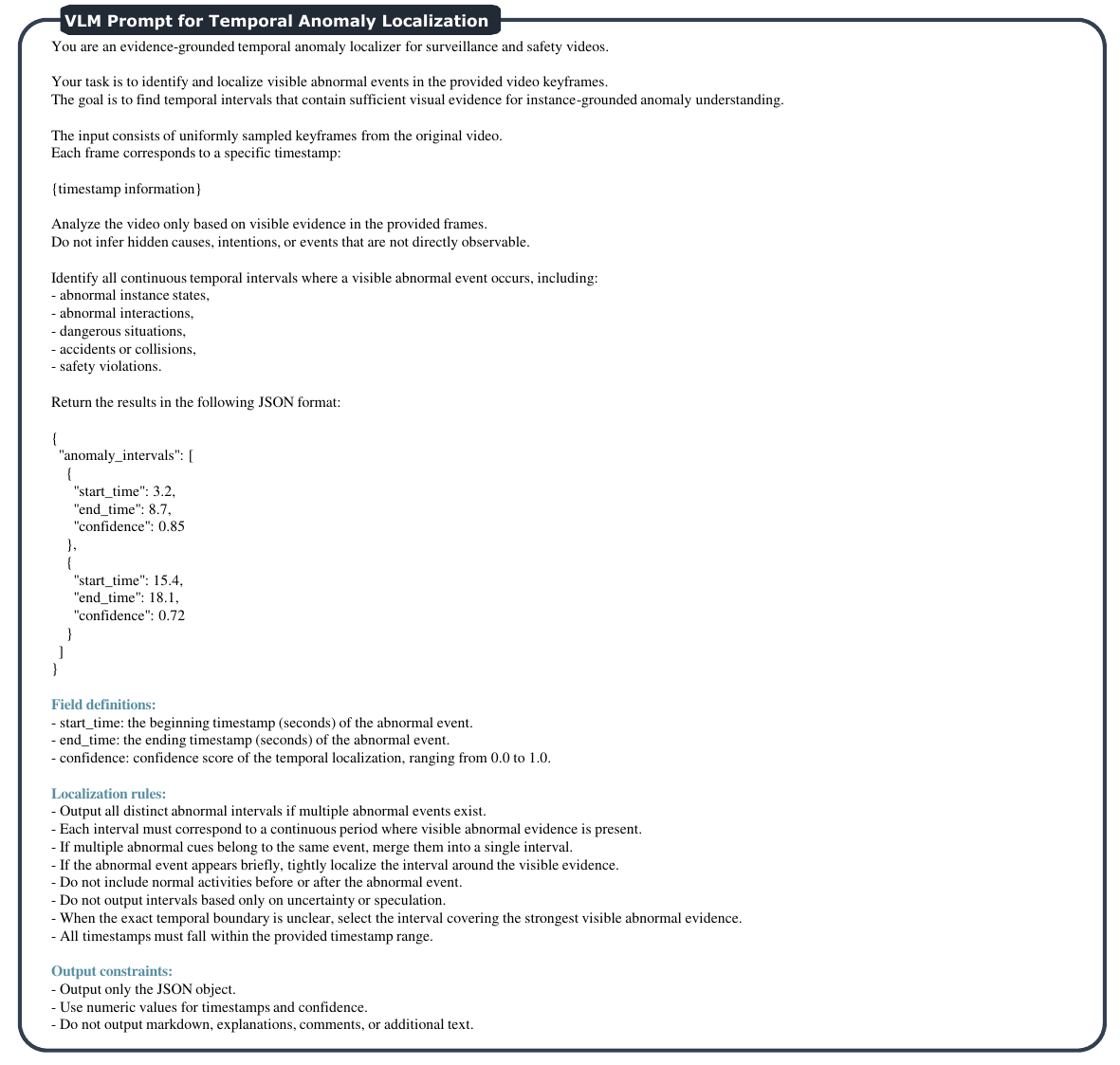}
    \caption{Prompt template for temporal anomaly localization. The model returns timestamped abnormal intervals in a structured JSON format.}
    \label{fig:tiou_eval_prompt}
\end{figure*}

The model receives sampled keyframes with timestamps and returns continuous intervals containing visible abnormal evidence. The prompt is shown in Fig.~\ref{fig:tiou_eval_prompt}.

The required output is a JSON object containing a list of intervals,
\(\texttt{anomaly\_intervals}=\{(s_k,e_k,\gamma_k)\}_{k=1}^{K}\),

where $s_k,e_k$ are start/end times and $\gamma_k$ is confidence. Timestamps are clipped to the video duration; invalid intervals receive zero tIoU. If multiple intervals are returned, the one with highest overlap to the reference is evaluated.

\subsection{\airs{} Evaluation}

\begin{figure*}[t]
    \centering
    \includegraphics[width=0.95\textwidth]{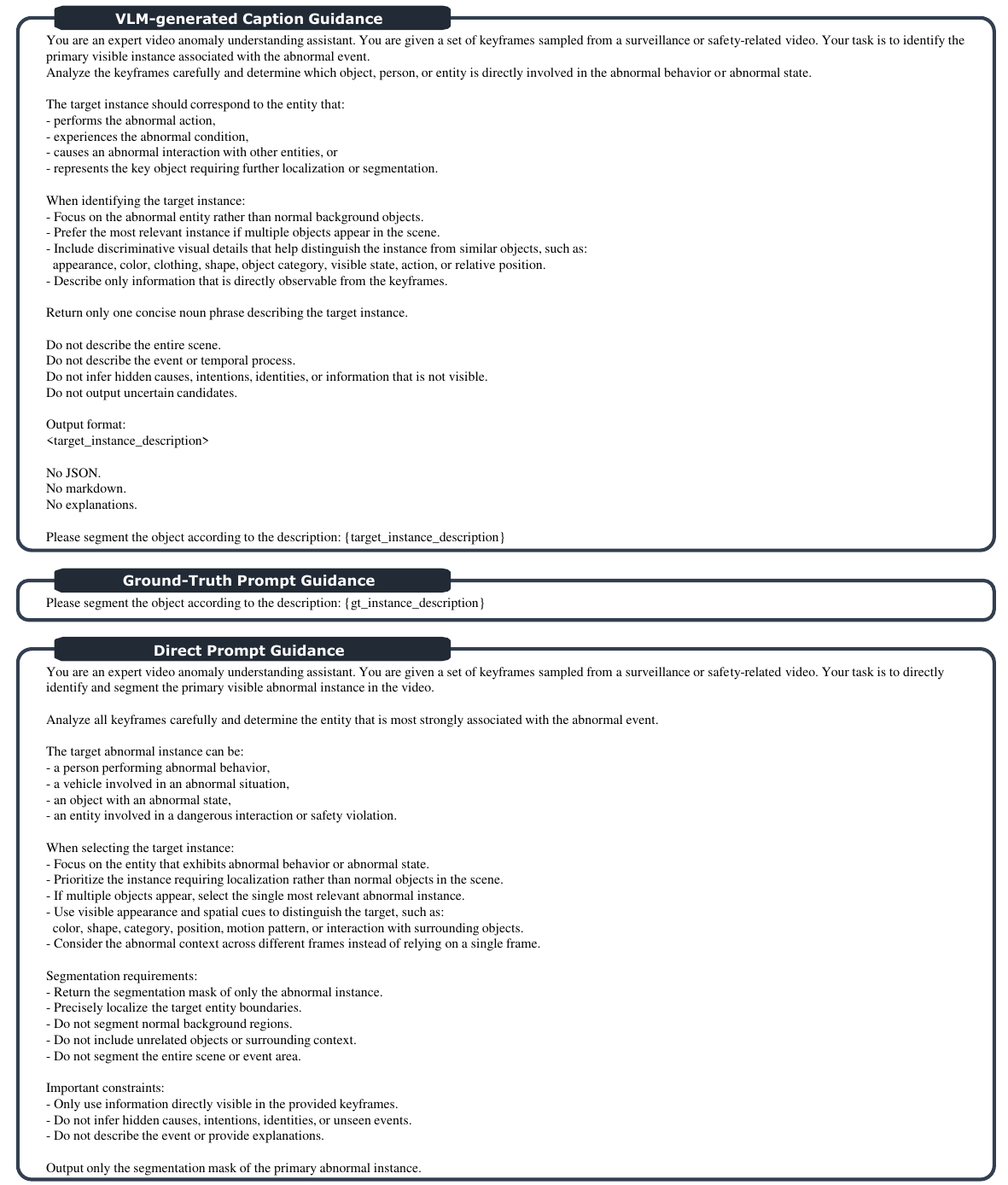}
    \caption{Prompt templates for \airs{} evaluation. The protocol includes direct prompting, generated-caption guidance, and ground-truth-caption guidance to disentangle semantic target selection from visual grounding.}
    \label{fig:airs_eval_prompt}
\end{figure*}

\airs{} evaluates mask-track prediction for the primary visible anomaly instance within the annotated anomaly interval. We use three prompt settings (Fig.~\ref{fig:airs_eval_prompt}) to separate semantic target selection from visual grounding.

\paragraph{Direct Prompt.}
The model directly identifies and segments the primary abnormal instance, testing end-to-end reasoning segmentation.

\paragraph{Generated-Caption Prompt.}
The model-generated target phrase is used as the segmentation query, testing semantic-to-spatial transfer.

\paragraph{GT-Caption Prompt.}
The ground-truth target phrase is used as the query, isolating visual grounding under correct semantic guidance.

\begin{figure*}[t]
    \centering
    \includegraphics[width=1\textwidth]{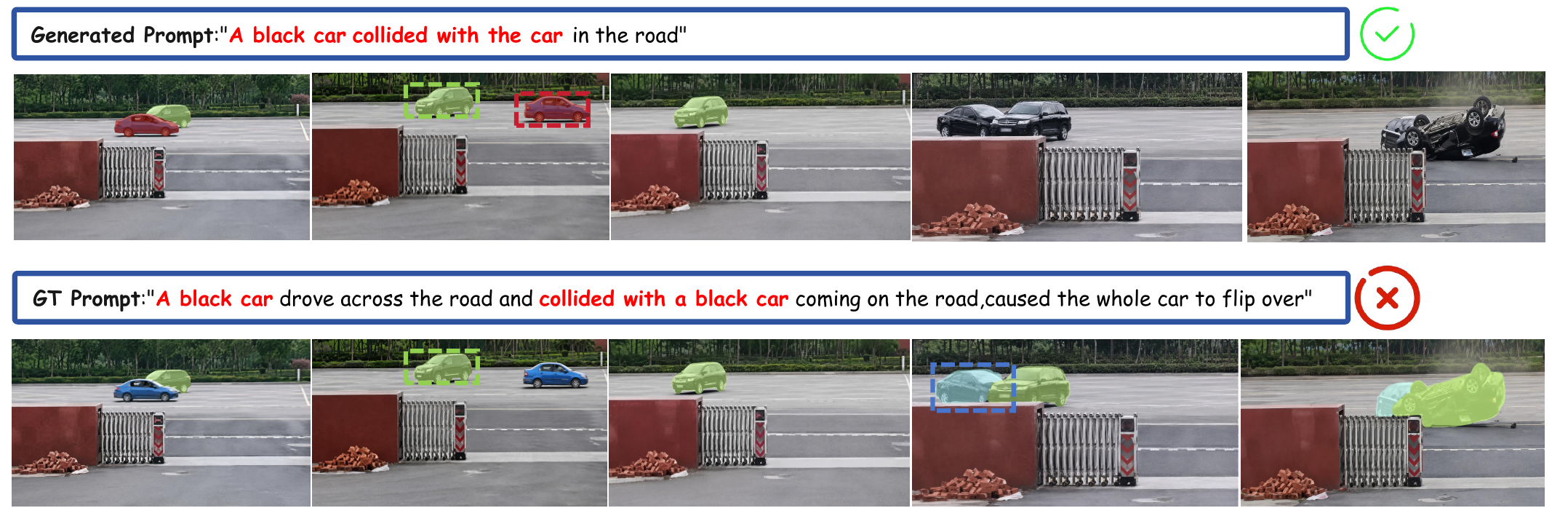}
    \caption{Generated-caption and GT-caption guidance for \airs{} using \textbf{Sa2VA-InternVL3-14B}. The generated prompt focuses on the anomalous interaction and produces a cleaner target track, whereas the more detailed GT prompt introduces distractor objects and causes incorrect mask propagation.}
    \label{fig:generated_gt}
\end{figure*}

\begin{figure*}[t]
    \centering
    \includegraphics[width=1\textwidth]{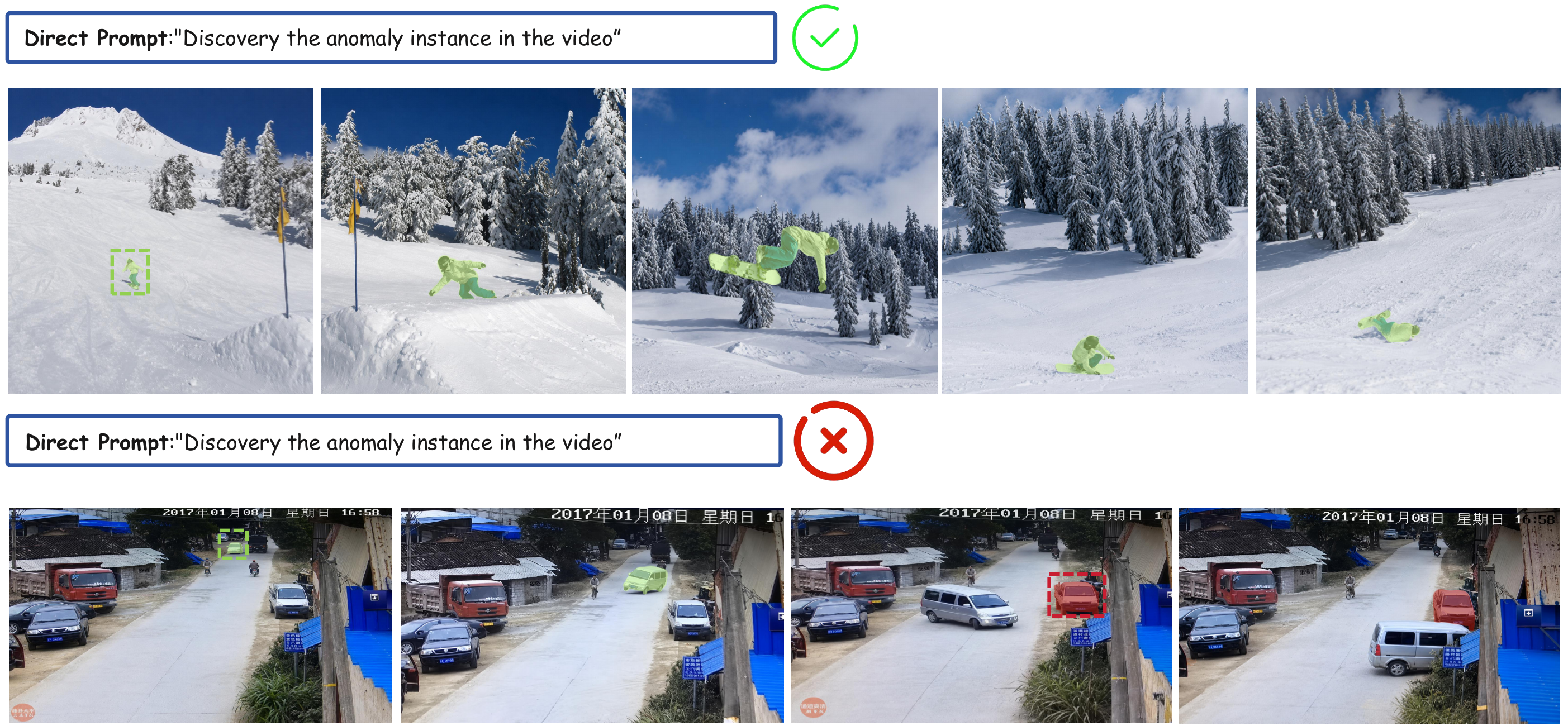}
    \caption{Direct-prompt \airs{} predictions by \textbf{Sa2VA-InternVL3-14B}. The model localizes visually clear anomaly instances, but can drift to salient distractors when the abnormal target is ambiguous or visually entangled with other objects.}
    \label{fig:direct_prompt}
\end{figure*}

\paragraph{Qualitative analysis.}
Figs.~\ref{fig:generated_gt} and \ref{fig:direct_prompt} illustrate the diagnostic value of the three \airs{} settings. Generated-caption and GT-caption guidance reveal whether errors come from the semantic query or mask propagation, while direct prompting evaluates end-to-end anomaly reasoning segmentation without an explicit text intermediate.

\subsection{Output Parsing and Failure Handling}

We deterministically parse all model outputs. For JSON-based tasks, invalid or unparsable video-level outputs count as negative predictions, while invalid temporal-localization outputs receive zero tIoU. Timestamps are clipped to video duration, and intervals with $e_k \le s_k$ are discarded. For A-IRS, empty or full-frame masks and predictions without a valid track within the annotated anomaly interval receive zero $J\&F$. Missing frames are treated as empty masks, making the final score reflect both spatial accuracy and track completeness.

\section{Metric Details}
\label{sec:supp_metric_details}
\subsection{Tracking Metrics}

\textbf{Video-level anomaly detection} is evaluated with anomaly-class recall:
\begin{equation}
\mathrm{Recall} = \frac{\mathrm{TP}}{\mathrm{TP} + \mathrm{FN}}.
\end{equation}

\textbf{Temporal anomaly localization} is evaluated with temporal intersection-over-union:
\begin{equation}
\mathrm{tIoU} =
\frac{|[s_p, e_p] \cap [s_g, e_g]|}
{|[s_p, e_p] \cup [s_g, e_g]|},
\end{equation}
where $[s_p,e_p]$ is the predicted interval and $[s_g,e_g]$ is the ground-truth interval.

\textbf{\airs{}} is evaluated within annotated anomaly intervals using region similarity $\mathcal{J}$, boundary accuracy $\mathcal{F}$, and their average $\mathcal{J}\&\mathcal{F}$.
Given a predicted binary mask $P_t$ and the ground-truth mask $G_t$ at frame $t$, the region similarity is defined as the intersection-over-union:
\begin{equation}
\mathcal{J}_t =
\frac{|P_t \cap G_t|}
{|P_t \cup G_t|}.
\end{equation}
The boundary accuracy $\mathcal{F}_t$ measures the F-measure between the predicted and ground-truth object boundaries. Let $B(P_t)$ and $B(G_t)$ denote the boundary pixel sets extracted from $P_t$ and $G_t$, and let matches be counted within a boundary tolerance radius $\delta$. Boundary precision and recall are
\begin{equation}
P^{\partial}_t =
\frac{
|\{p \in B(P_t): \exists g \in B(G_t), \|p-g\|_2 \leq \delta\}|
}
{|B(P_t)|},
\end{equation}
\begin{equation}
R^{\partial}_t =
\frac{
|\{g \in B(G_t): \exists p \in B(P_t), \|g-p\|_2 \leq \delta\}|
}
{|B(G_t)|}.
\end{equation}
The boundary F-measure is then
\begin{equation}
\mathcal{F}_t =
\frac{2P^{\partial}_t R^{\partial}_t}
{P^{\partial}_t + R^{\partial}_t}.
\end{equation}
For an annotated anomaly interval $\mathcal{T}$, the final track-level scores are averaged over frames:
\begin{equation}
\mathcal{J} =
\frac{1}{|\mathcal{T}|}
\sum_{t\in\mathcal{T}} \mathcal{J}_t,
\quad
\mathcal{F} =
\frac{1}{|\mathcal{T}|}
\sum_{t\in\mathcal{T}} \mathcal{F}_t,
\end{equation}
and
\begin{equation}
\mathcal{J}\&\mathcal{F}
=
\frac{\mathcal{J}+\mathcal{F}}{2}.
\end{equation}

\subsection{Understanding Metric: TraceScore}
\label{sec:tracescore-validation}

\paragraph{Motivation.}
Standard text similarity metrics can score anomaly captions highly when lexically related yet grounded to the wrong instance, temporal process, location, or scene-level risk. TraceScore therefore combines a VLM semantic judge, a lexical-support penalty, and a contradiction gate to evaluate video-grounded semantic consistency. We instantiate the VLM judge with Qwen3.5-122B for all TraceScore computations.

\begin{figure*}[t]
    \centering
    \includegraphics[width=0.98\textwidth]{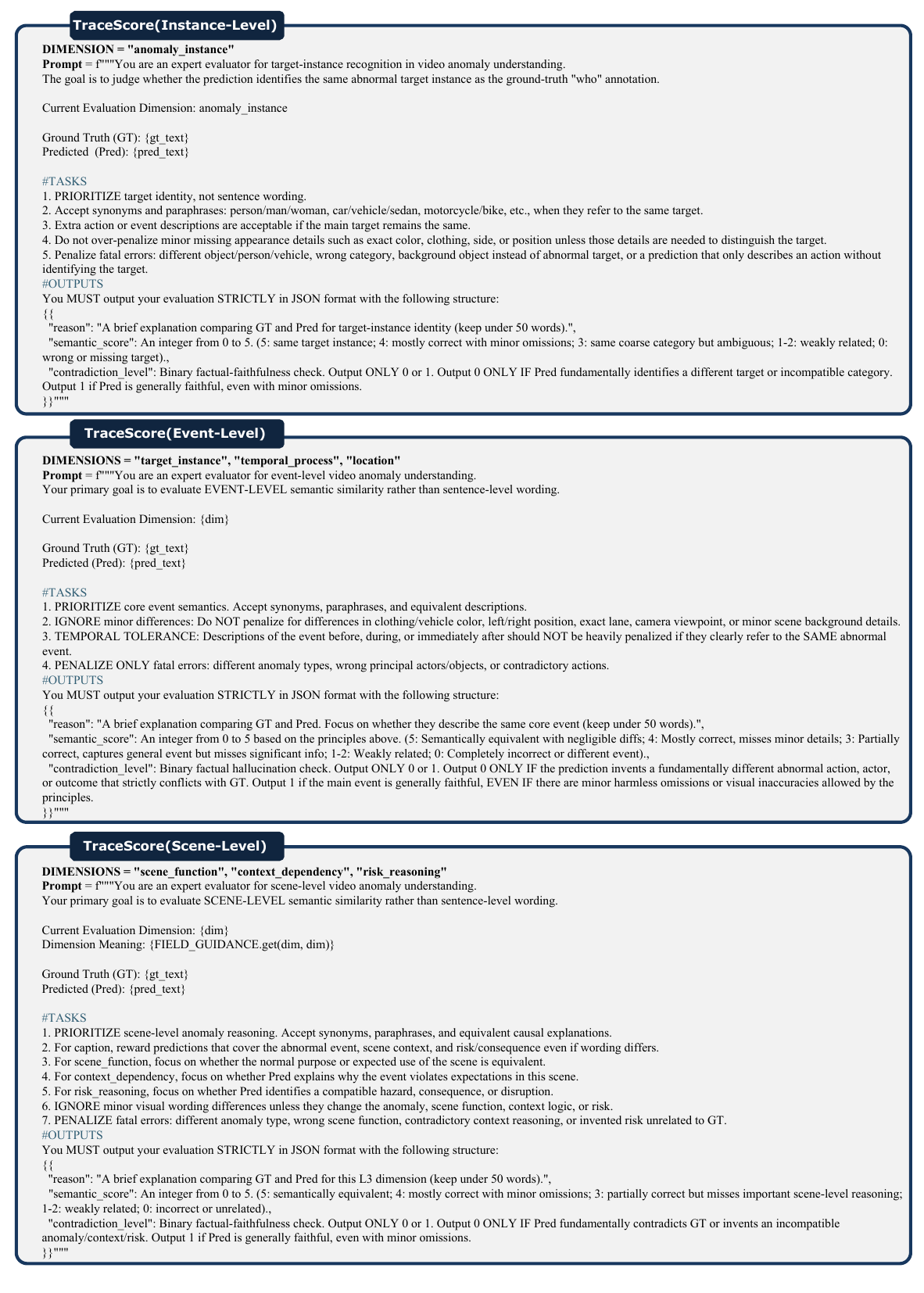}
    \caption{Prompt template for \tscore{} evaluation. The VLM judge scores semantic consistency at instance, event, and scene levels while checking contradictions against the reference annotation and video-grounded anomaly semantics.}
    \label{fig:tracescore_prompt}
\end{figure*}

For sample $i$ and semantic field $d$, let $c_{i,d}$ and $\hat c_{i,d}$ denote the reference and predicted captions. The VLM judge returns a normalized semantic similarity score $s^{\mathrm{sem}}_{i,d}$, and METEOR provides a lexical-support score $s^{\mathrm{lex}}_{i,d}$. We define
\begin{align}
p_{i,d} &= \max(0, s^{\mathrm{sem}}_{i,d} - s^{\mathrm{lex}}_{i,d}), \\
\mathrm{TraceScore}_{i,d}
&= C_{i,d} \cdot s^{\mathrm{sem}}_{i,d} \cdot (1 - \lambda p_{i,d}),
\end{align}
where $C_{i,d}\in\{0,1\}$ is a contradiction gate. If the prediction contradicts the reference or the video evidence, $C_{i,d}=0$; otherwise, $C_{i,d}=1$. The final score is averaged over all annotated semantic fields:
\begin{equation}
\mathrm{TraceScore}_{i}
= \frac{1}{|\mathcal{D}_i|}
\sum_{d\in\mathcal{D}_i}
\mathrm{TraceScore}_{i,d}.
\end{equation}
The complete judge prompt is provided in Fig.~\ref{fig:tracescore_prompt}.

\paragraph{Human calibration of $\lambda$.}
We calibrate $\lambda$ on an independent 200-video subset spanning different sources, scenes, camera types, anomaly categories, and prediction qualities. Annotators inspect the video, focal track, reference field, and model prediction, then assign each field a consistency label: inconsistent, partially consistent, or consistent, while also marking explicit contradictions. We encode the ordinal labels as 0, 1, and 2.Let \(h_{i,d}^{(a)}\) denote annotator \(a\)'s label for field \(d\)
of sample \(i\). For each sample, we first average the ordinal labels
across annotators for each semantic field, and then average the
field-level scores to obtain the video-level human consistency score
\(\bar h_i\).

We search $\lambda\in[0,1]$ with a coarse-to-fine window search and maximize Spearman correlation with $\bar{h}_{i}$:
\begin{equation}
R(\lambda)
= \rho_{\mathrm{Spearman}}
\left(
\mathrm{TraceScore}_{i}(\lambda),
\bar{h}_{i}
\right),
\end{equation}
The selected $\lambda=0.38$ is fixed for all experiments. Results are reported in Table~\ref{tab:lambda-calibration}.

\begin{table}[t]
\centering
\caption{Calibration of the lexical-support penalty $\lambda$ on the 200-video human-labeled subset.}
\label{tab:lambda-calibration}
\resizebox{\columnwidth}{!}{
\begin{tabular}{lccc}
\toprule
Search range & Step size & Best $\lambda$ & Spearman $\rho$ \\
\midrule
$[0,1]$ & \textbf{0.1} & \textbf{0.4} & \textbf{0.746} \\
\textbf{[0.3,0.4]} & \textbf{0.01} & \textbf{0.38} & \textbf{0.752} \\
Final & -- & \textbf{0.38} & \textbf{0.752} \\
\bottomrule
\end{tabular}}
\end{table}

\paragraph{Correlation with human judgments.}
After fixing $\lambda$, we compare TraceScore with lexical metrics, embedding-based metrics, and a VLM judge without the penalty/gate on a held-out set disjoint from the 200-video calibration subset. For metric $m$, let $z_i^{(m)}$ be its score for sample $i$. All metrics use the same field aggregation and are compared with $\bar{h}_i$:
\begin{align}
r^{(m)}
&=
\operatorname{Corr}_{\mathrm{P}}
\left(
z_i^{(m)},\bar{h}_i
\right), \\
\rho^{(m)}
&=
\operatorname{Corr}_{\mathrm{P}}
\left(
\operatorname{rank}(z_i^{(m)}),
\operatorname{rank}(\bar{h}_i)
\right), \\
\tau_b^{(m)}
&=
\frac{N_c-N_d}
{\sqrt{(N_c+N_d+T_m)(N_c+N_d+T_h)}},
\end{align}
where $N_c$ and $N_d$ are concordant and discordant sample pairs, and $T_m,T_h$ account for ties. Correlations are reported in Table~\ref{tab:tracescore-human-correlation}.

\begin{table}[t]
\centering
\caption{Correlation with human semantic-consistency judgments. TraceScore better reflects human judgments.}
\label{tab:tracescore-human-correlation}
\resizebox{\columnwidth}{!}{
\begin{tabular}{lccc}
\toprule
Metric & Pearson $r$ & Spearman $\rho$ & Kendall $\tau$ \\
\midrule
BLEU-4 & 0.421   &      0.438    &      0.352 \\
METEOR & 0.536   &      0.551     &     0.443 \\
BERTScore & 0.561   &      0.574     &     0.461 \\
TraceScore w/o contradiction gate & 0.712   &      0.726     &     0.604 \\
TraceScore & \textbf{0.768}   &      \textbf{0.781}      &    \textbf{0.659} \\
\bottomrule
\end{tabular}}
\vspace{-7pt}
\end{table}

\begin{figure*}[t]
    \centering
    \includegraphics[width=0.98\textwidth]{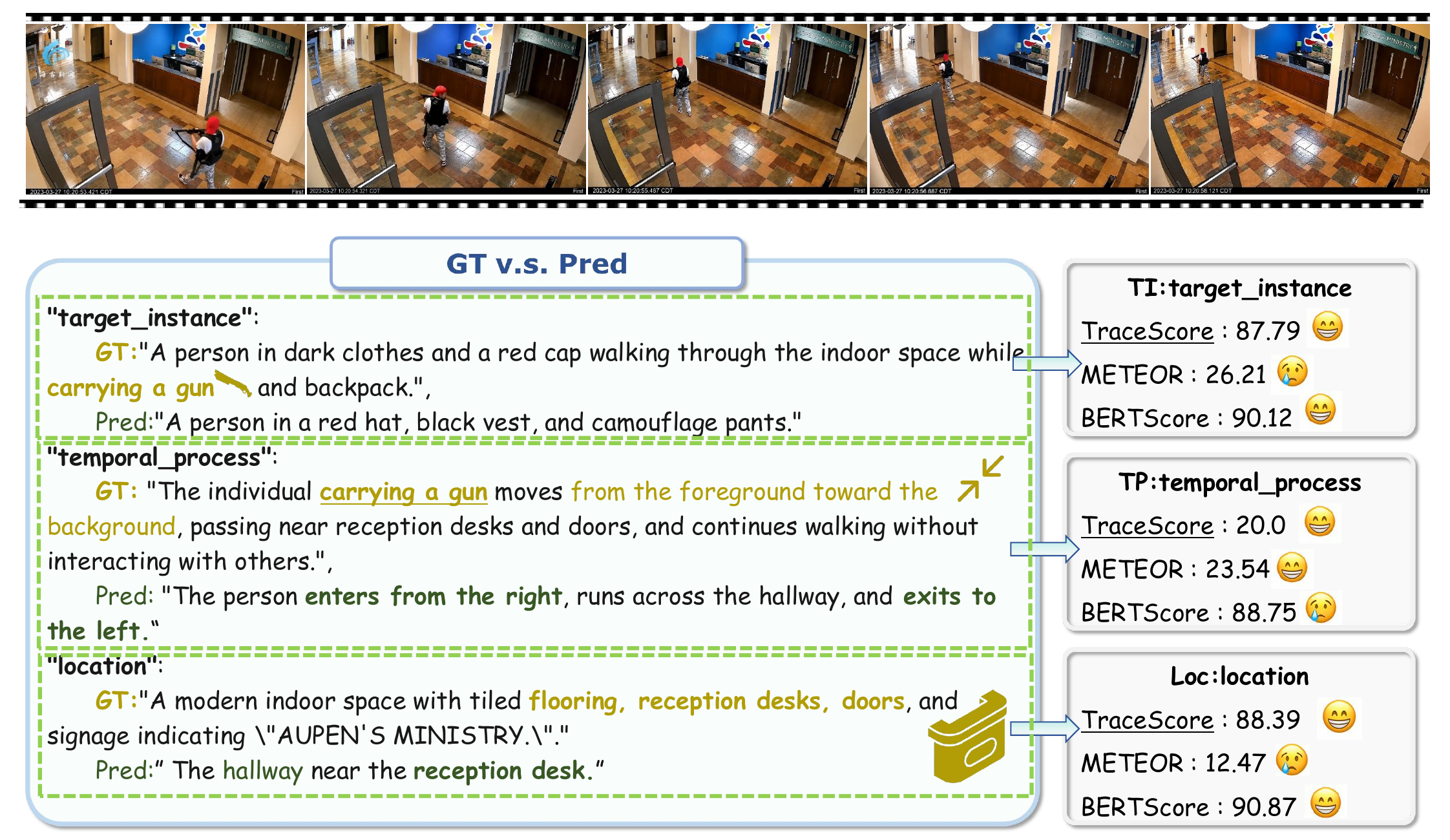}
    \caption{Event-level \tscore{} case on \textbf{InternVL3.5-14B} outputs. Compared with METEOR and BERTScore, \tscore{} better penalizes predictions that are lexically or semantically close but miss the target instance, temporal process, or event location.}
    \label{fig:trace3}
\end{figure*}

\begin{figure*}[t]
    \centering
    \includegraphics[width=0.98\textwidth]{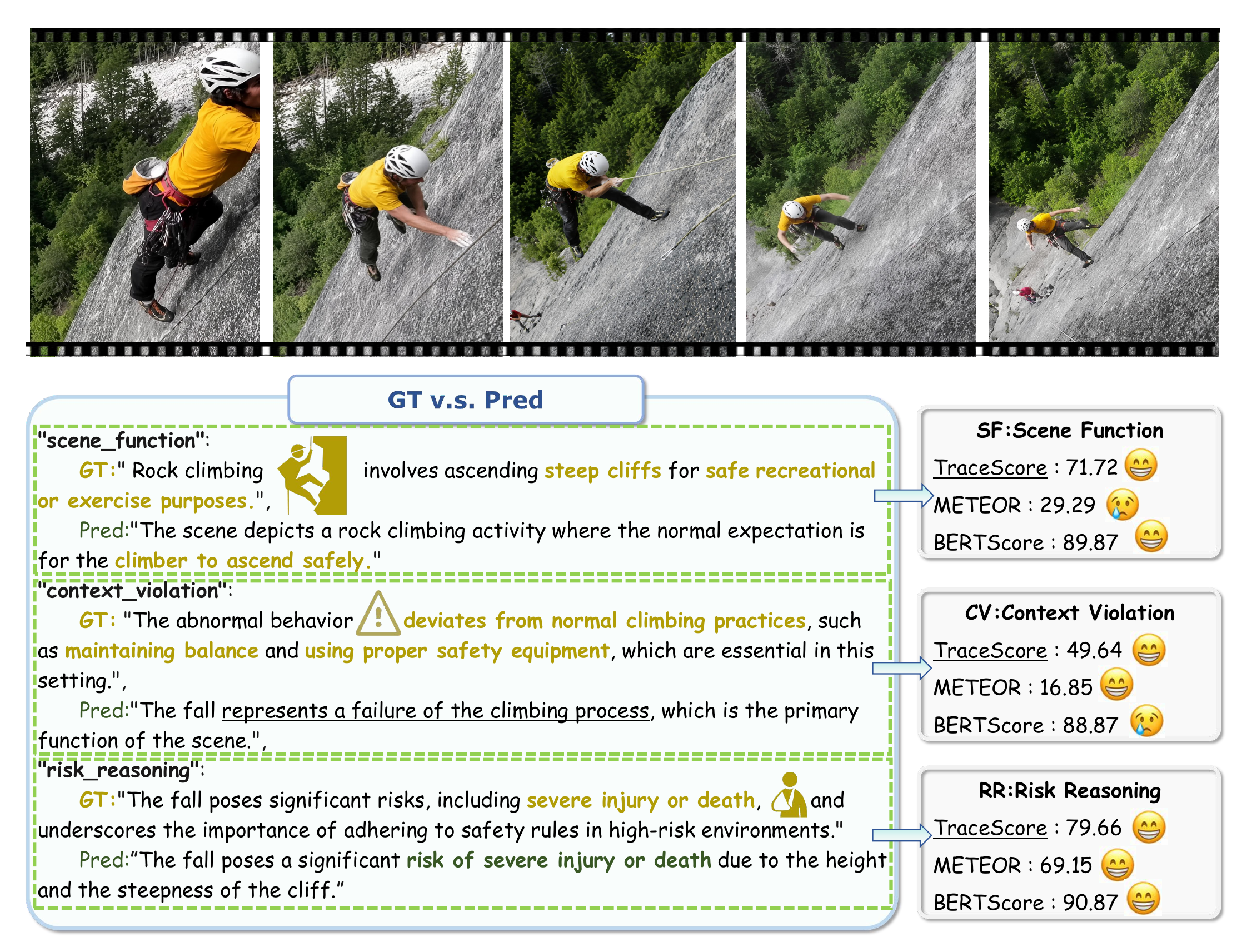}
    \caption{Scene-level \tscore{} case on \textbf{InternVL3.5-14B} outputs. Compared with METEOR and BERTScore, \tscore{} better penalizes predictions that are superficially related but inconsistent with the scene function, context violation, or risk reasoning.}
    \label{fig:trace1}
\end{figure*}

\paragraph{Qualitative case analysis.}
Fig.~\ref{fig:trace3} and Fig.\ref{fig:trace1} compare text-similarity metrics with \tscore{}, showing that \tscore{} better penalizes grounding-inconsistent predictions and is more suitable for fine-grained anomaly understanding.

\begin{figure*}[t]
    \centering
    \includegraphics[width=1\textwidth]{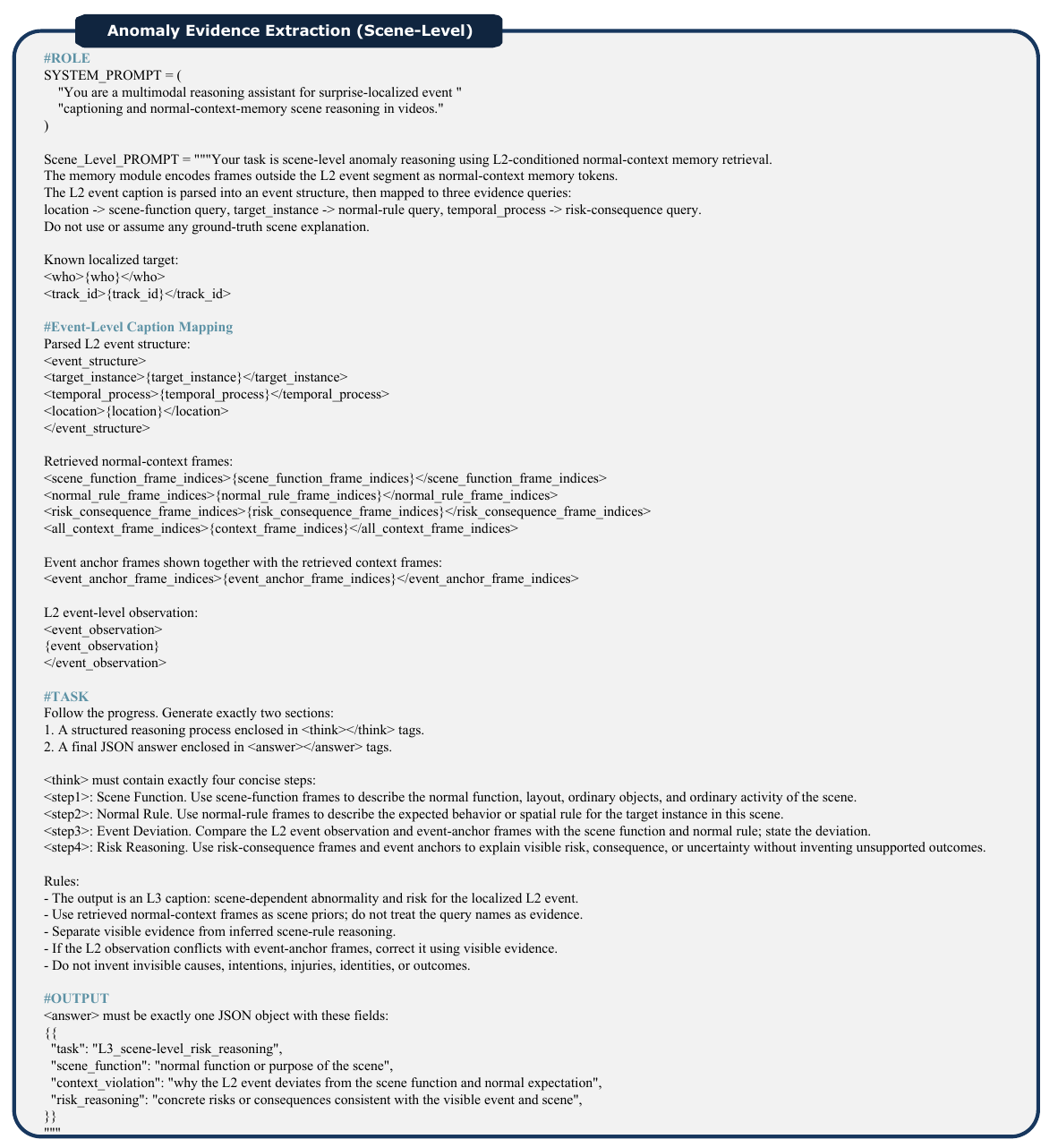}
    \caption{Prompt template for scene-level reasoning, which uses L2-conditioned normal-context retrieval to generate scene function, context violation, and risk reasoning.}
    \label{fig:l3_prompt}
\end{figure*}



